\documentclass{article}
\usepackage{graphicx,times,amsthm,amsfonts,amsmath,amscd,amssymb,algorithm,algorithmic}
\usepackage[dvips]{color}
\newtheorem{dfn}{Definition}
\newtheorem{thm}{Theorem}
\newtheorem{lemma}{Lemma}

\newtheorem{prop}{Proposition}%

\newcommand{\whv}{\widehat{v}} 
\renewcommand{\P}{\mathbf{P}}
\newcommand{\E}{\mathbf{E}}
\newcommand{\bfs}{\mathrm{bfs}}
\newcommand{\child}{\textrm{child}}
\newcommand{\parent}{\textrm{parent}}
\newcommand{\flt}{\lfloor t\rfloor}

\begin{document}
\title{Rumors in a Network: Who's the Culprit?}
\author{Devavrat~Shah~and~Tauhid~Zaman}
\maketitle


\begin{abstract}
We provide a systematic study of the problem of finding the source of a rumor in a network.  We model rumor spreading in a network with a variant of the popular SIR model and then construct an estimator for the rumor source.  This estimator is based upon a novel topological quantity which we term \textbf{rumor centrality}.  We establish that this is an ML estimator for a class of graphs.  We find the following surprising threshold phenomenon: on trees which grow faster than a line, the estimator always has non-trivial detection probability, whereas on trees that grow like a line, the detection probability will go to 0 as the network grows.  Simulations performed on synthetic networks such as the popular small-world and scale-free networks, and on real networks such as an internet AS network and the U.S. electric power grid network, show that the estimator either finds the source exactly or within a few hops of the true source across different network topologies.  We compare rumor centrality to another common network centrality notion known as distance centrality.  We prove that on trees, the rumor center and distance center are equivalent, but on general networks, they may differ.  Indeed, simulations show that rumor centrality outperforms distance centrality in finding rumor sources in networks which are not tree-like.
\end{abstract}



%

\section{Introduction}

In the modern world the ubiquity of networks has 
made us vulnerable to new types of network risks. These network 
risks arise in many different contexts, but share a common structure: 
an isolated risk is amplified because it is spread by the network.  
For example, as we have witnessed in the recent financial crisis, 
the strong dependencies or `network' between institutions have 
led to the situation where the failure of one (or few) institution(s) 
have led to global instabilities. In an electrical power grid
network, an isolated failure could lead to a rolling blackout. Computer
viruses utilize the Internet to infect millions of computers everyday.
Finally, malicious rumors or misinformation can rapidly spread through
existing social networks and lead to pernicious effects on individuals
or society. 
In all of these situations, a policy maker, power network operator, 
Internet service provider or victim of a malicious rumor, 
would like to identify the source of the risk as quickly as 
possible and subsequently quarantine its effect. 

In essence, all of these situations can be modeled as a rumor 
spreading through a network, where the goal is to find the source 
of the rumor in order to control and prevent these 
network risks based on limited information about the network
structure and the `rumor infected' nodes.  In this paper, we shall
take important initial steps towards a systematic study of the 
question of identifying the rumor source based on the network 
structure and rumor infected nodes, as well as understand the 
fundamental limitations on this estimation problem.

\subsection{Related Work}

Prior work on rumor spreading has primarily focused on viral 
epidemics in populations. The natural (and somewhat standard) 
model for viral epidemics is known as the \emph{susceptible-infected-recovered} 
or SIR model \cite{ref:sir}. In this model, there are three types 
of nodes: (i) susceptible nodes, capable of being infected;
(ii) infected nodes that can spread the virus further; and 
(iii) recovered nodes that are cured and can no longer become 
infected.  Research in the SIR model has focused on understanding 
how the structure of the network and rates of infection/cure lead
to large epidemics \cite{ref:epidemic-sw},\cite{ref:epidemic-sf},\cite{ref:newman},\cite{ref:topology}. This 
motivated various researchers to propose network inference techniques
for learning the relevant network parameters \cite{ref:mcmc1},\cite{ref:mcmc2},\cite{ref:mcmc3},\cite{ref:mcmc4},\cite{ref:nhpp}.  
However, there has been little (or no) work done on inferring the 
source of an epidemic. 

The primary reason for the lack of such work is that it is quite 
challenging. To substantiate this, we briefly describe a closely 
related (and much simpler)  problem of reconstruction on trees 
\cite{ref:evans},\cite{ref:mossel}, or more generally, on graphs 
\cite{ref:montanari}.  In this problem one node in the graph, call 
it the {\em root} node, starts with a value, say $0$ or $1$. 
This information is propagated to its neighbors and their neighbors 
recursively along a breadth-first-search (BFS) tree of the graph 
(when the graph is a tree, the BFS tree is the graph). Now each 
transmission from a node to its neighbor is noisy -- a transmitted 
bit is flipped with a small probability. The question of interest 
is to estimate or reconstruct the value of the root node,
based on the `noisy' information received at nodes that are far 
away from root. Currently, this problem is well understood only 
for graphs that are trees or tree-like, after a long history. 
Now the rumor source identification problem is, in a sense harder, 
as we wish to identify the location of the source among many nodes 
based on the infected nodes -- clearly a much noisier situation than 
the reconstruction problem. Therefore, as the first step, we would 
like to understand this problem on trees.  

\subsection{Our Contributions}

In this paper, we take initial steps towards understanding the 
question of identifying the rumor source in a network based on 
(rumor) infected nodes. Specifically, we start by considering
a probabilistic model of rumor spreading in the network as the
ground truth. This model is based on the SIR model which is
well studied in the context of epidemiology, as mentioned 
earlier. It is the natural rumor spreading model 
with minimal side information. Therefore, such a model
provides the perfect starting point to undertake the systematic 
study of such inference problems.  

The question of interest is to identify the source of the rumor
based on information about the infected nodes as well
as the underlying network structure using the prior information
about the probabilistic rumor spreading model. In the absence
of additional information (i.e. a uniform prior), clearly the maximum
likelihood (ML) estimator minimizes the estimation error. 
Therefore, we would like to identify (a) a computationally 
tractable representation of the ML estimator if possible, and 
(b) evaluate the detection probability of such an estimator. 

Now obtaining a succinct, useful characterization of an ML estimator 
for a general graph seems intractable. Therefore, following the 
philosophical approach of researchers working on the reconstruction 
problem mentioned above and on the efficient graphical model based 
inference algorithm (i.e. Belief Propagtation), we address the 
above questions for tree networks. 

We are able to obtain a succinct and computationally efficient
characterization of the ML estimator for the rumor source when
the underlying network is a regular tree. We are able to 
characterize the correct detection probability of this ML
estimator for regular trees of any given node degree $d$. 
We find the following phase transition. For $d = 2$, i.e. when
the network is a linear graph (or a path), the asymptotic detection 
probability is $0$. For $d \geq 3$, i.e. when the network is
an {\em expanding} tree, the asymptotic detection probability is
strictly positive. For example, for $d = 3$, we identify it to
be $1/4$. 

As the next step, we consider non-regular trees. The ML estimator
of regular trees naturally extends to provide a rumor source estimator
for non-regular trees. However, it is not necessarily the ML estimator.
We find that when a non-regular tree satisfies a certain 
{\em geometric} growth property (see \ref{ssec:geom} for the precise 
definition), then the asymptotic detection probability of this
estimator is $1$. This suggests that even though this computationally
simple estimator is not the ML estimator, its asymptotic performance 
is as good as any other (and hence the ML) estimator. 


Motivated by results for trees, we develop a natural, computationally
efficient heuristic estimator for general graphs based on the ML estimator
for regular trees. We perform extensive simulations to show that this 
estimator performs quite well on a broad range of network topologies. 
This includes synthetic networks obtained from the small-world model and 
the scale-free model as well as real network
topologies such as the U.S. electric power-grid and the Internet.   
In summary, we find that when the network structure (irrespective of
being a tree) is not too {\em irregular}, the estimator performs well.



The estimator, which is ML for regular trees, can be thought of as
assigning a non-negative value to each node in a tree. We call this
value the \emph{rumor centrality} of the node. In essence, the estimator
chooses the node with the highest rumor centrality as the estimated
source, which we call the {\em rumor center} of the network. There are
various notions of network centralities that are popular in the
literature (cf. \cite{ref:dc},\cite{ref:bc}). Therefore, in principle, each of these
network centrality notions can act as rumor source estimators. 
Somewhat surprisingly, we find that the source estimator based on the
popular {\em distance centrality} notion is identical to the rumor
centrality based estimator for any tree. Therefore, in a sense our
work provides theoretical justification for distance centrality in 
the context of rumor source detection. 

Technically, the method for establishing non-trivial asymptotic 
detection for regular trees with $d\geq 3$ is quite 
different from that for geometric trees. Specifically, for 
regular trees with $d \geq 3$, we need to develop 
a refined probabilistic estimation of the rumor spreading process 
to establish our results. Roughly speaking, this is necessary
because the rumor process exhibits high variance on
expanding trees (due to exponential growth in the size of the neighborhood
of a node with distance) and hence standard concentration based results are 
not meaningful (for establishing the result).  On 
the other hand, for geometric trees the rumor process 
exhibits sharp enough concentration (due to sub-exponential
growth in the size of the neighborhood of a node with distance)
for establishing the desired result. This also allows
us to deal with heterogeneity in the context of geometric trees. 
Similar technical contrasts between geometric and 
expanding structures are faced in analyzing growth processes 
on them. For example, in the classical percolation literature 
precise `shape theorems' are known for
geometric structures (e.g. $d$-dimensional grids) \cite{ref:shape1,ref:shape2,ref:shape3,ref:shape4,ref:shape5, ref:shape6}.  However, little is known in the context of expanding structures. 
Indeed, our techniques 
for analyzing regular (expanding) trees do overcome 
such challenges. Developing them further for general
expanding graphs (non-regular trees and beyond) remain an
important direction for future research. 

Finally, we note that calculating the rumor centrality of a 
node is equal to computing the number of possible linear extensions 
of a given partial order represented by the tree structure rooted 
at that particular node. Subsequently, our algorithm leads to the
fastest known algorithm for computing the number of possible
linear extensions in this context (see \cite{ref:posetalg} for the
best known algorithm in the literature). 

\subsection{Organization}

In Section \ref{sec:estimator}, the probabilistic model for rumor 
spreading and the derivation of the source estimator is presented. 
Section \ref{sec:rcfacts} studies properties of this estimator and
presents an efficient algorithm for its evaluation. Section 
\ref{sec:mainres} presents results about the effectiveness of 
the estimator for tree networks in terms of its asymptotic 
detection probability. Section \ref{sec:simulations} shows 
the effectiveness of the estimator for general networks by 
means of extensive simulations. Section \ref{sec:proofs} provides
detailed proofs of the result presented in Section \ref{sec:mainres}. 
We conclude in Section \ref{sec:conclusion} with directions for future
work.

\section{Rumor Source Estimator}\label{sec:estimator}

In this section we start with a description of our rumor 
spreading model and then we define the maximum likelihood (ML) estimator 
for the rumor source.  For regular tree graphs, we equate the ML
estimator to a novel topological quantity which we call rumor centrality.  We then use rumor centrality to construct rumor source estimators for general graphs.

\subsection{Rumor Spreading Model}\label{ssec:model}
We consider a network of nodes modeled as an undirected graph $G(V,E)$, 
where $V$ is a countably infinite set of nodes and $E$ is the set of 
edges of the form $(i,j)$ for some $i$ and $j$ in $V$.  We assume 
the set of nodes is countably infinite in order to avoid boundary 
effects. We consider the case where initially only one node 
$v^*$ is the rumor source.  

We use a variant of the common SIR model for the rumor spreading 
known as the \emph{susceptible-infected} or SI model which does not 
allow for any nodes to recover, i.e. once a node has the rumor, it 
keeps it forever. Once a node $i$ has the rumor, it is able to spread 
it to another node $j$ if and only if there is an edge between them, i
.e. if $\left(i,j\right)\in E$.  Let $\tau_{ij}$ be the time it
takes for node $j$ to receive the rumor from node $i$ once $i$ has
the rumor. In this model, $\tau_{ij}$ are independent and have 
exponential distribution with parameter (rate) $\lambda$.  Without 
loss of generality, assume $\lambda=1$.


\subsection{Rumor Source Estimator: Maximum Likelihood (ML)}\label{ssec:ml}

Let us suppose that the rumor starting at a node, say $v^*$ at time $0$ has spread
in the network $G$. We observe the network at some time and find $N$ infected
nodes. By definition, these nodes must form a connected subgraph of $G$. We shall
denote it by $G_N$. Our goal is to produce an estimate, which we shall denote by 
$\whv$, of the original source $v^*$ based on the observation $G_N$ and the 
knowledge of $G$. To make this estimation, we know that the rumor has spread
in $G_N$ as per the SI model described above. However, a priori we do not know
from which source the rumor started. Therefore, we shall assume a uniform prior
probability of the source node among all nodes of $G_N$. With respect to this
setup, the maximum likelihood (ML) estimator of $v^*$ with respect to the SI
model given $G_N$ minimizes the error probability, i.e. maximizes the correct 
detection probability. By definition, the ML estimator is 
\begin{align}
		\whv &\in \arg \max_{v \in G_N} \mathbf{P}(G_N|v),  
\end{align}
where $\P(G_N|v)$ is the probability of observing $G_N$ under the SI model assuming
$v$ is the source, $v^*$. Thus, ideally we would like to evaluate $\P(G_N|v)$ for
all $v \in G_N$ and then select the one with the maximal value (ties broken
uniformly at random).


\subsection{Rumor Source Estimator: ML for Regular Trees}\label{sec:regtree}

In general, evaluation of $\P(G_N|v)$ may not be computationally tractable. 
Here we shall show that for regular trees, $\P(G_N|v)$ becomes proportional 
to a quantity $R(v,G_N)$ which we define later and call rumor centrality. 
The $R(v,G_N)$ is a topological quantity and is intimately related to the
structure of $G_N$.  

\begin{figure}[t]
	\centering
		\includegraphics{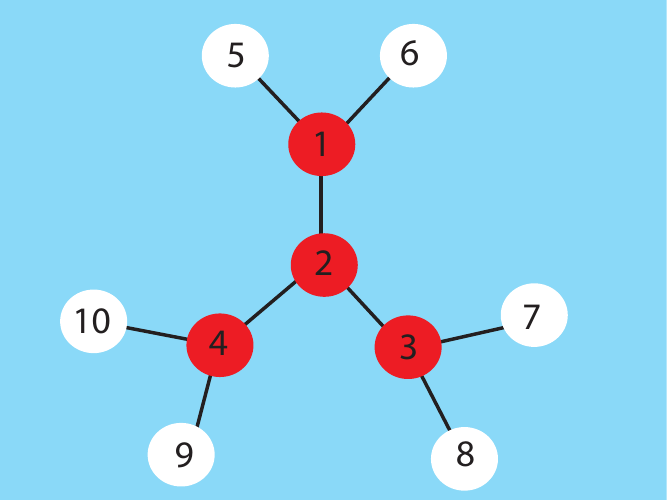}
		\caption{Example network where the rumor graph has four nodes.}  
	\label{fig:regexample}
\end{figure}

Now to evaluate $\P(G_N|v)$ when the underlying graph is a tree, essentially 
we wish to find the probability of all possible events that result in
$G_N$ after $N$ nodes are infected starting with $v$ as the source under
the SI model. To understand such events, let us consider a simple example 
as shown  in Figure \ref{fig:regexample} with $N = 4$. Now, suppose node 1 
was the source, i.e. we wish to calculate $\P(G_4|1)$. Then there are two
disjoint events or node orders in which the rumor spreads that will 
lead to $G_4$ with $1$ as the source: $\big\{1, 2,3, 4\big\}$ and 
$\big\{1,2,4,3 \big\}$. However, due to the structure of the network,
infection order $\big\{1,3,2,4\big\}$ is not possible. Therefore, in 
general to evaluate $\P(G_N|v)$, we need to find all such 
{\em permitted permutations} and their corresponding probabilities. 

Let $\Omega(v,G_N)$ be the set of all permitted permutations starting 
with node $v$ and resulting in rumor graph $G_N$. We wish to determine the
probability $\P(\sigma|v)$ for each $\sigma \in \Omega(v, G_N)$. To 
that end, let $\sigma = \big\{v_1 = v, v_2,\dots, v_N\big\}$. Let
us define, $G_k(\sigma)$ as the subgraph (of $G_N$) containing
nodes $\big\{v_1 = v, v_2,\dots, v_k\big\}$ for $1\leq k\leq N$. Then,
\newcommand{\bsep}{\;\big|\;}
\begin{align}\label{eq:l1}
\P\big(\sigma \bsep v\big) & = \prod_{k=2}^{N} \P\big(k^{\mbox{{\small th}}} \mbox{~infected node}=v_{k} \;\big|\; G_{k-1}(\sigma),v \big).
\end{align}

Each term in the product on the right hand side in \eqref{eq:l1}, can be
evaluated as follows. Given $G_{k-1}(\sigma)$ (and source $v$), the next
infected node could be any of the neighbors of nodes in $G_{k-1}(\sigma)$
which are not yet infected. For example, in Figure \ref{fig:regexample} $G_2$ is
$\{1,2\}$ when the source is assumed to be $1$. In that case, the next 
infected node could be any one of the $4$ nodes: $3,4,5$ and $6$. Now 
due to the memoryless property of exponential random variables and since
all infection times on all edges are independent and identically
distributed (i.i.d.), it follows that each of these nodes is equally likely
to be the next infected node. Therefore, each one of them has probability
$1/4$. More generally, if $G_{k-1}(\sigma)$ has $n_{k-1}(\sigma)$ uninfected neighboring 
nodes, then each one of them is equally likely to be the next infected node
with probability $1/n_{k-1}(\sigma)$. Therefore, \eqref{eq:l1} reduces
to
\begin{align}\label{eq:l2}
\P\big(\sigma \bsep v\big) & = \prod_{k=2}^{N} \frac{1}{n_{k-1}(\sigma)}. 
\end{align}
Given \eqref{eq:l2}, now the problem of computing $\P\big(\sigma \bsep v\big)$
boils down to evaluating the size of the {\em rumor boundary} $n_{k-1}(\sigma)$
for $2\leq k\leq N$. To that end, suppose the $k^{th}$ added node to $G_{k-1}(\sigma)$
is $v_k(\sigma)$ with degree $d_k(\sigma)$. Then it contributes $d_{k}(\sigma)-2$ new edges (and hence nodes
in the tree) to the rumor boundary. This is because, $d_{k}(\sigma)$ new edges are added 
but we must remove the edge along which the recent infection happened, which is counted twice.  
That is, $n_{k}(\sigma) = n_{k-1}(\sigma)+d_{k}(\sigma)-2$. Subsequently, 

\begin{align}\label{eq:l3}
n_k(\sigma) & = d_{1}(\sigma) + \sum_{i=2}^{k} (d_{i}(\sigma)-2). 
\end{align}

Therefore, 

\begin{align}\label{eq:l4}
 \P\big(\sigma \bsep v\big) & = \prod_{k=2}^{N} \frac{1}{d_{1}(\sigma) + \sum_{i=2}^{k} (d_{i}(\sigma)-2)}. 
\end{align}

For a $d$ regular tree, since all nodes have the same degree $d$, it follows from \eqref{eq:l4}
that every permitted permutation $\sigma$ has the same probability, independent of the source.  
Specifically, for any source $v$ and permitted permutation $\sigma$
\begin{align*} 
\P\big(\sigma \bsep v\big) & = \prod_{k=1}^{N-1}\frac{1}{d k- 2 (k-1)}\\
													& \equiv p(d,N).
\end{align*}
From above, it follows immediately that for a $d$ regular tree, for
any $G_N$ and candidate source $v$, $\P\big(G_N \bsep v\big)$ is 
proportional to $|\Omega(v,G_N)|$. Formally, we shall denote
the number of distinct permitted permutations $|\Omega(v,G_N)|$
by $R(v,G_N)$.
\begin{dfn} Given a graph $G$ and vertex $v$ of $G$, we define 
$R(v,G)$ as the total number of distinct permitted permutations 
of nodes of $G$ that begin with node $v\in G$ and respect the
graph structure of $G$. 
\end{dfn} 
In summary, the ML estimator for a regular tree becomes
\begin{align}
	\widehat{v} &\in \arg \max_{v\in G_N} \P\big(G_N \bsep v \big) \nonumber\\
							&= \arg \max_{v\in G_N} \sum_{\sigma\in \Omega(v,G_N)}\P\big(\sigma \bsep v\big) \nonumber\\
							&=\arg \max_{v\in G_N} R(v,G_N) p(d,N)\nonumber\\
	            &= \arg \max_{v\in G_N}R(v,G_N)\label{eq:estimator}
\end{align}
with ties broken uniformly at random. %

\subsection{Rumor Source Estimator: General Trees}

As \eqref{eq:estimator} suggests, the ML estimator for a regular tree can
be obtained by simply evaluating $R(v,G_N)$ for all $v$. However, as indicated
by \eqref{eq:l4}, such is not the case for a general tree with heterogeneous degree. 
This is because in the regular tree, all permitted permutations were equally 
likely, whereas in a general tree, different permitted permutations have different 
probabilities. To form an ML estimator for a general tree we would need to keep 
track of the probability of every permitted permutation. This could be computationally
quite expensive  because of the exponential number of terms involved.  Therefore, 
we construct a simple heuristic to take into account the degree heterogeneity.

Our heuristic is based upon the following simple idea.  The likelihood of a 
node is a sum of the probability of every  permitted permutation for which 
it is the source.  In general, these will have different values, but it may 
be that a majority of them have a common value.  We then need to determine this 
value of the probability of the common permitted permutations.  To do this, we 
assume the nodes receive the rumor in a breadth-first search (BFS) fashion. 
Roughly speaking, this corresponds to the fastest or most probable spreading
of the rumor.

\begin{figure}[t]
	\centering
		\includegraphics{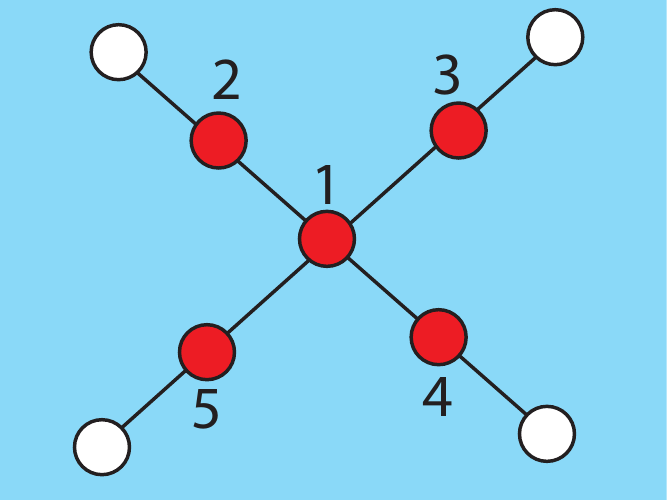}
		\caption{Example network where rumor centrality with the BFS heuristic equals the likelihood $\mathbf{P}(G_N|v)$.  The rumor infected nodes are in red and labeled with numbers.}  
	\label{fig:bfsprior}
\end{figure}

To calculate the BFS permitted permutation probability, we construct a 
sequence of nodes in a BFS fashion, with the source node fixed.  For example, 
consider the network in Figure \ref{fig:bfsprior}. If we let node $2$ be the 
source, then a BFS sequence of nodes would be ${2,1,3,4,5}$ and the probability of this permitted permutation is
given by \eqref{eq:l4}.

If we define the BFS permitted permutation with node $v$ as the source as $\sigma_v^*$, 
then the rumor source estimator becomes (ties broken uniformly at random)
\begin{align}\label{eq:estimator.1}
		\widehat{v} &\in \arg \max_{v\in G_N} \mathbf{P}\big(\sigma_v^* \bsep v\big) R(v,G_N). 
\end{align}
We now consider an example to show the effect of the BFS heuristic.  For the network in Figure \ref{fig:bfsprior}, the corresponding estimator 
value for node $1$ is

\begin{align*}
		 \mathbf{P}(\sigma_1^*)R(1,G_N)&= \left(\frac{1}{4}\right)^44!\\
		 														   &= 6\left(\frac{1}{4}\right)^3 
\end{align*}

and for node $2$ it is

\begin{align*}
		 \mathbf{P}(\sigma_2^*)R(2,G_N)&= \frac{1}{2}\left(\frac{1}{4}\right)^33!\\
		 														   &= 3\left(\frac{1}{4}\right)^3. 
\end{align*}

For comparison, the exact likelihood of node $1$ is

\begin{align*}
		 \mathbf{P}(G_N|1)=& \mathbf{P}(\{1,2,3,4,5\}|1)+\mathbf{P}(\{1,2,3,5,4\}|1)\\
		                      &+\mathbf{P}(\{1,2,4,3,5\}|1)+\dots	+\mathbf{P}(\{1,5,4,3,2\}|1)\\
		                      =& 24\left(\frac{1}{4}\right)^4\\ 
		 											=& 6\left(\frac{1}{4}\right)^3  
\end{align*}

and for node $2$ it is

\begin{align*}
		 \mathbf{P}(G_N|2)=&    \mathbf{P}(\{2,1,3,4,5\}|2)+\mathbf{P}(\{2,1,3,5,4\}|2)\\
		                      &+\mathbf{P}(\{2,1,4,3,5\}|2)+\mathbf{P}(\{2,1,4,5,3\}|2)\\ 			
		                      &+\mathbf{P}(\{2,1,5,3,4\}|2)+\mathbf{P}(\{2,1,5,4,3\}|2)\\
		                      =& 6\frac{1}{2}\left(\frac{1}{4}\right)^3\\ 
		 											=& 3\left(\frac{1}{4}\right)^3.  
\end{align*}

In this case we find that the BFS heuristic equals the true likelihood for both nodes.  Second, node $1$ is only twice as likely as node $2$ to be the source.  However, if we look at the ratio of the rumor centralities of the nodes we find 
\begin{align*}
	\frac{R(1,G_N)}{R(2,G_N)} & = \frac{4!}{3!}\\
						&= 4.
\end{align*}

Thus, the rumor centrality of node $1$ is four times as large as that of node $2$.  What is happening is 
that without the BFS heuristic, rumor centrality
 is being fooled to always select higher degree nodes because it assumes all nodes have the same 
 degree.  Therefore, if a node only has a few infected neighbors (such as node $2$), rumor 
 centrality assumes that the node was not immediately infected and consequently did not have 
 time to infect its neighbors.  However, the BFS heuristic tries to compensate for the 
 tendency of rumor centrality to favor higher degree nodes.  
 
 Indeed, as we shall see in Section \ref{sec:simulations}, this heuristic is an improvement over
the naive extension of the estimator \eqref{eq:estimator} for networks with very heterogeneous degree 
distributions. That is, biasing as per $\mathbf{P}\big(\sigma_v^* \bsep v\big)$ in \eqref{eq:estimator.1} is better than the unbiased version of it. 
 
\subsection{Rumor Source Estimator: General Graphs}

The ML estimator for a general graph, in principle can be computed by
following a similar approach as that for general trees. Specifically, 
it corresponds to computing the summation of the likelihoods of all
possible {\em permitted permutations} given the network structure. 
This could be computationally prohibitive. Therefore, we propose a
simple heuristic. 

To that end, note that even in a general graph the rumor spreads along 
a spanning tree of the observed graph corresponding to the {\em first time} 
each node receives the rumor.  Therefore, a reasonable approximation 
for computing the likelihood $\P(G_N|v)$ is as follows. First, suppose
we know which spanning tree was involved in the rumor spreading. Then, 
using this spanning tree, we could apply the previously developed tree 
estimator.  However, it is the lack of knowledge of the spanning tree 
that makes the rumor source estimation problem complicated.  

We circumvent the issue of not knowing the underlying spanning tree as
follows. We assume that if node $v\in G_N$ was the source, then the 
rumor spreads along a breadth first search (BFS) tree rooted at $v$, 
$T_{\bfs}(v)$.  The intuition is that if $v$ was the source, then the BFS 
tree would correspond to the fastest (intuitively, most
likely) spread of the rumor. Therefore, effectively we obtain the 
following rumor source estimator for a general rumor graph $G_N$:
\begin{align}\label{eq:estimator.2}
  \widehat{v} & \in \arg\max_{v\in G_N} \mathbf{P}\big(\sigma_v^* \bsep v\big) R(v,T_{\bfs}(v)).
\end{align}
In the above ties are broken uniformly at random as before. Also, like in \eqref{eq:estimator.1}, 
the $\sigma_v^*$ represents the BFS ordering of nodes in the tree $T_{\bfs}(v)$.

\begin{figure}[t]
	\centering
		\includegraphics{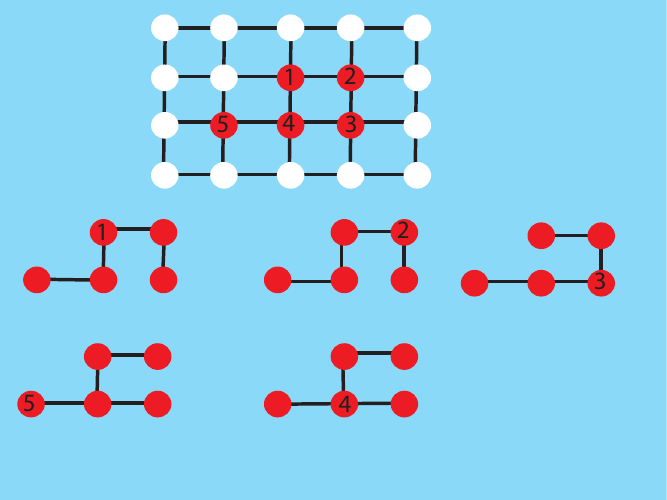}
		\caption{Example network with a BFS tree for each node shown.  The rumor infected nodes are shown in red.}  
	\label{fig:rcbfs}
\end{figure}

For example, consider the network in Figure \ref{fig:rcbfs}.  The BFS trees for each node are shown.  Using the expression for $R(v,G_N)$  from Section \ref{ssec:rc}, the general graph estimator values for the nodes are
\begin{align*}
	\mathbf{P}\big(\sigma_1^* \bsep 1\big) R(1,T_{\bfs}(1)) & = \frac{1}{4*6*8*10}\frac{5!}{20}\\
	\mathbf{P}\big(\sigma_2^* \bsep 2\big) R(2,T_{\bfs}(2)) & = \frac{1}{4*6*8*10}\frac{5!}{30}\\
	\mathbf{P}\big(\sigma_3^* \bsep 3\big) R(3,T_{\bfs}(3)) & = \frac{1}{4*6*8*10}\frac{5!}{20}\\
	\mathbf{P}\big(\sigma_4^* \bsep 4\big) R(4,T_{\bfs}(4)) & = \frac{1}{4*6*8*10}\frac{5!}{10}\\
	\mathbf{P}\big(\sigma_5^* \bsep 5\big) R(5,T_{\bfs}(5)) & = \frac{1}{4*6*8*10}\frac{5!}{40}.
\end{align*}

Node $4$ maximizes this value and would be the estimate of the rumor source for this network.  We will show with simulations that this general graph estimator performs well on different network topologies.

\section{Rumor Centrality: Properties \& Algorithm}\label{sec:rcfacts}

The quantity $R(v,G_N)$ plays an important role in each of the rumor
source estimators \eqref{eq:estimator}, \eqref{eq:estimator.1} and 
\eqref{eq:estimator.2}. Recall that $R(v,G_N)$ counts the number of
distinct ways a rumor can spread in the network $G_N$ starting from source 
$v$. Thus, it assigns each node of $G_N$ a non-negative number or score.
We shall call this number,  $R(v,G_N)$, the \textbf{rumor centrality} 
of the node $v$ with respect to $G_N$. The node with maximum rumor centrality
will be called the \textbf{rumor center} of the network.  Indeed, the rumor 
center is the ML estimation of the rumor source for regular trees. 

This section describes ways to evaluate $R(v,G_N)$ efficiently when $G_N$ is a tree. It
also describes an important property of rumor centrality that will be useful
in establishing our main results later in the paper. Further, we discuss a
surprising relation between the rumor center and the so called distance center
of a tree. Finally, we remark on the relation between rumor centrality and
the number of linear extensions of a partially ordered set described by a
tree graph. 

\subsection{Rumor Centrality: Succinct Representation}\label{ssec:rc}

\begin{figure}[t]
	\centering
		\includegraphics{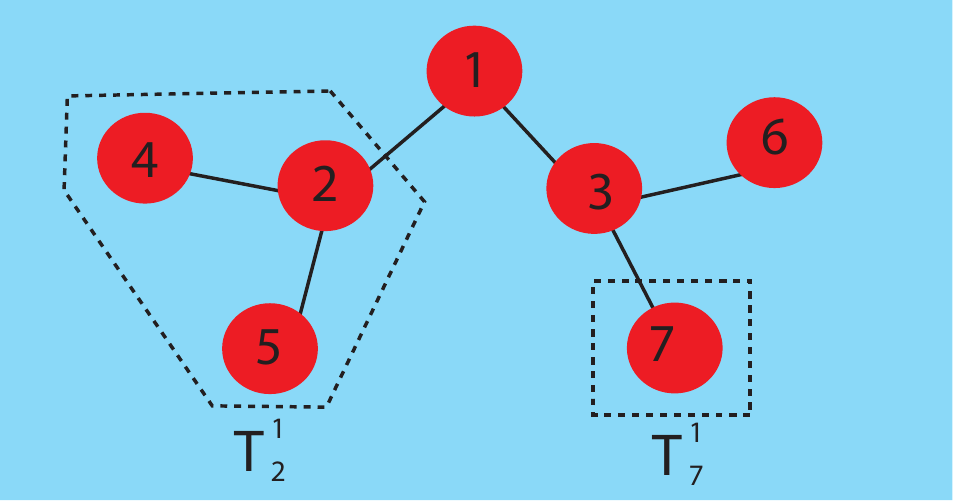}
		\caption{Illustration of subtree variable $T_{u}^v$.}  
	\label{fig:subtree}
\end{figure}

Let $G_N$ be a tree graph. Define $T_{u}^v$ as the number of nodes in 
the subtree rooted at node $u$, with node $v$ as the source.  To illustrate 
this notation, a simple example is shown in Figure \ref{fig:subtree}.  
Here $T_2^1 = 3$ because there are 3 nodes in the subtree with node 2 as 
the root and node 1 as the source.  Similarly, $T_7^1 =1$ because 
there is only 1 node in the subtree with node 7 as the root and node 1 as the source.  

We now can count the permitted permutations of $G_N$ with $v$ as the source.  
In the following analysis, we will abuse notation and use $T_{u}^v$ to refer 
to both the subtrees and the number of nodes in the subtrees. Recall that we
are looking for permitted permutations of $N$ nodes of $G_N$. That is, 
we have $N$ slots in a given permitted permutation, the first of which must 
be the source node $v$. The question is, how many distinct ways can we 
fill the remaining $N-1$ slots. The basic constraint is due to the
causality induced by the tree graph that a node $u$ must come before
all the nodes in its subtree $T_u^v$. Given a slot assignment for all
nodes in $T_u^v$ subject to this constraint, there are $R(u,T_u^v)$ 
different ways in which these nodes can be ordered. This suggests a natural 
recursive relation between the rumor centrality $R(v,G_N)$ and the
rumor centrality of its immediate children's subtrees $R(u, T_u^v)$ with
$u \in \child(v)$. Here $\child(v)$ represents the set of all children
of $v$ in tree $G_N$ assuming $v$ as its root. Specifically, there is
no constraint between the orderings of the nodes of different subtrees $T_u^v$ 
with $u \in \child(v)$. This leads to the following relation. 

\begin{align}\label{eq:r1}
	R(v,G_N) &=(N-1)!\prod_{u\in \child(v)}\frac{R(u,T_u^v)}{T_u^v!}.
\end{align}

To understand the above expression, note that the number of ways to 
partition $N-1$ slots for different subtrees is 
$(N-1)!\prod_{u\in \child(v)}\frac{1}{T_u^v!}$,  and the partition 
corresponding to $T_u^v, ~u \in \child(v)$ leads to $R(u,T_u^v)$
distinct orderings, thus resulting in \eqref{eq:r1}. 

If we expand this recursion \eqref{eq:r1} to the next level of 
depth in $G_N$ we obtain	
\begin{align*}
	R(v,G_N) &=(N-1)!\prod_{u\in \child(v)}\frac{R(u,T_u^v)}{T_u^v!}\\
		&= (N-1)!\prod_{u\in \child(v)}\frac{(T_u^v-1)!}{T_u^v!}\prod_{w\in \child(u)}\frac{R(w,T_w^v)}{T_w^v!}\\
		&=(N-1)!\prod_{u\in \child(v)}\frac{1}{T_u^v}\prod_{w\in \child(u)}\frac{R(w,T_w^v)}{T_w^v!}.\\
\end{align*}
A leaf node $l$ will have have 1 node and 1 permitted permutation, so $R(l,T_l^v)=1$. 
If we continue this recursion until we reach the leaves of the tree, then we find that 
the number of permitted permutations for a given tree $G_N$ rooted at $v$ is 
\begin{align}
	R(v,G_N)&=(N-1)!\displaystyle\prod_{u \in G_N \backslash v}\frac{1}{ T_{u}^v}\nonumber\\
	       &= N!\displaystyle\prod_{u \in G_N}\frac{1}{ T_{u}^v}.
\end{align}
In the last line, we have used the fact that $T_v^v = N$.  We thus end up with a simple expression for rumor centrality in terms of the size of the subtrees of all nodes in $G_N$.

\begin{figure}[t]
	\centering
		\includegraphics{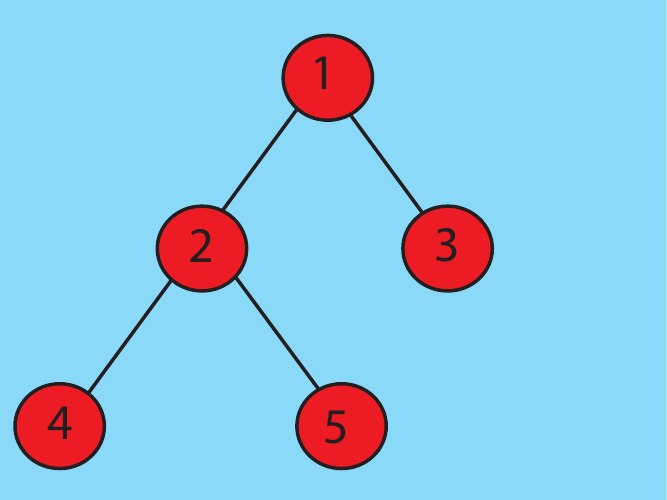}
		\caption{Example network for calculating rumor centrality.}  
	\label{fig:rcexample}
\end{figure}

As an example of the use of rumor centrality, consider the network in Figure \ref{fig:rcexample}.  Using the rumor centrality formula, we find that the rumor centrality of node 1 is
\begin{align*}
	R(1,G) &= \frac{5!}{5*3}=8.
\end{align*}
Indeed, there are 8 permitted permutations of this network with node 1 as the source, which we list below.
\begin{align*}
  &\{1,3,2,4,5\},\{1,2,3,4,5\},\{1,2,4,3,5\},\{1,2,4,5,3\},\\	
  &\{1,3,2,5,4\},\{1,2,3,5,4\},\{1,2,5,3,4\},	\{1,2,5,4,3\}.
\end{align*}

\subsection{Rumor Centrality via Message-Passing}\label{ssec:algorithm}

In order to find the rumor center of an $N$ node tree $G_N$, 
we need to first find the rumor centrality of every node in $G_N$.  
To do this we need the size of the subtrees $T_{u}^v$ for all $v$ and $u$ 
in $G_N$.  There are $N^2$ of these subtrees. Therefore, a naive algorithm
can lead to $\Omega(N^2)$ operations. We shall utilize a local relation between
the rumor centrality of neighboring nodes in order to calculate it in 
$O(N)$ computation in a distributed, message-passing manner.
    
To this end, consider two neighboring nodes $u$ and $v$ in $G_N$.  
All of their subtrees will be the same size except for those rooted at $u$ and 
$v$.  In fact, there is a special relation between these two subtrees.
  \begin{equation}
  T_{u}^{v} = N- T_{v}^{u}. 
  \end{equation}
For example, in Figure \ref{fig:subtree}, for node 1, $T_2^1$ has 3 nodes, 
while for node 2, $T_1^2$ has $N-T_2^1$ or 4 nodes.  Because of this relation, 
we can relate the rumor centralities of any two neighboring nodes.
  \begin{equation}
  R(u,G_N) = R(v,G_N) \frac{T^{v}_{u}}{N- T^{v}_{u}}.\label{eq:algoR} 
  \end{equation}  
This result is the key to our algorithm for calculating the rumor 
centrality for all nodes in $G_N$.  We first select any node $v$ 
as the source node and calculate the size of all of its subtrees $T_{u}^{v}$ 
and its rumor centrality $R(v,G_N)$.  This can be done by having each node 
$u$ pass two messages up to its parent.  The first message is the number of 
nodes in $u$'s subtree, which we call $t^{up}_{u\rightarrow \parent(u)}$.  
The second message is the cumulative product of the size of the subtrees of 
all nodes in $u$'s subtree, which we call $p^{up}_{u\rightarrow \parent(u)}$.  
The parent node then adds the $t^{up}_{u\rightarrow \parent(u)}$ messages 
together to obtain the size of its own subtree, and multiplies the 
$p^{up}_{u\rightarrow \parent(u)}$  messages together to obtain its 
cumulative subtree product.  These messages are then passed upward 
until the source node receives the messages.  By multiplying the 
cumulative subtree products of its children, the source node will 
obtain its rumor centrality, $R(v,G_N)$.  

With the rumor centrality of node $v$, we then evaluate the rumor 
centrality for the children of $v$ using equation \eqref{eq:algoR}.  
Each node passes its rumor centrality to its children in a message 
we define as $r^{down}_{u\rightarrow u'}$ for $u' \in \child(u)$. 
Each node $u$ can calculate its rumor centrality using its 
parent's rumor centrality and its own subtree size $T_u^{v}$. 
We recall that the rumor centrality of a node is the number 
of permitted permutations that result in $G_N$.  Thus, this 
message-passing algorithm is able to count the (exponential) 
number of permitted permutations for every node in $G_N$ using only 
$O(N)$ computations.  The pseudocode for this message-passing algorithm 
is included for completeness.
\begin{algorithm}
\caption{Rumor Centrality Message-Passing Algorithm} \label{alg:rc}
\begin{algorithmic} [1]
\STATE Choose a root node $v\in G_N$ 
\FOR{$u$ in $G_N$} 
	\IF{u is a leaf}
		\STATE $t^{up}_{u\rightarrow \parent(u)} = 1$
		\STATE $p^{up}_{u\rightarrow \parent(u)} = 1$
	\ELSE
		\IF{u is root v}
			\STATE {$\forall~v' \in \child(v)$:}
			$r^{down}_{v\rightarrow v'} = \frac{N!}{\displaystyle N\prod_{j \in \child(v)}                                         p^{up}_{j\rightarrow v}} $ 
	\ELSE
		\STATE $t^{up}_{u\rightarrow \parent(u)} = \displaystyle\sum_{j\in \child(u)} t^{up}_{j\rightarrow u} +1$
		\STATE $p^{up}_{u\rightarrow \parent(u)} =  t^{up}_{u\rightarrow \parent(u)} \displaystyle\prod_{j\in \child(u)} p^{up}_{j\rightarrow u} $
 		\STATE $\forall~u' \in \child(u)$:~$r^{down}_{u\rightarrow u'} = r^{down}_{\parent(u)\rightarrow u}                                                   \frac{t^{up}_{u\rightarrow \parent(u)}}{N-t^{up}_{u\rightarrow \parent(u)}}$
 		\ENDIF
 		\ENDIF
 		\ENDFOR
\end{algorithmic}
\end{algorithm}

\subsection{A Property of Rumor Centrality}\label{rc_prop}
\begin{figure}[t]
	\centering
		\includegraphics{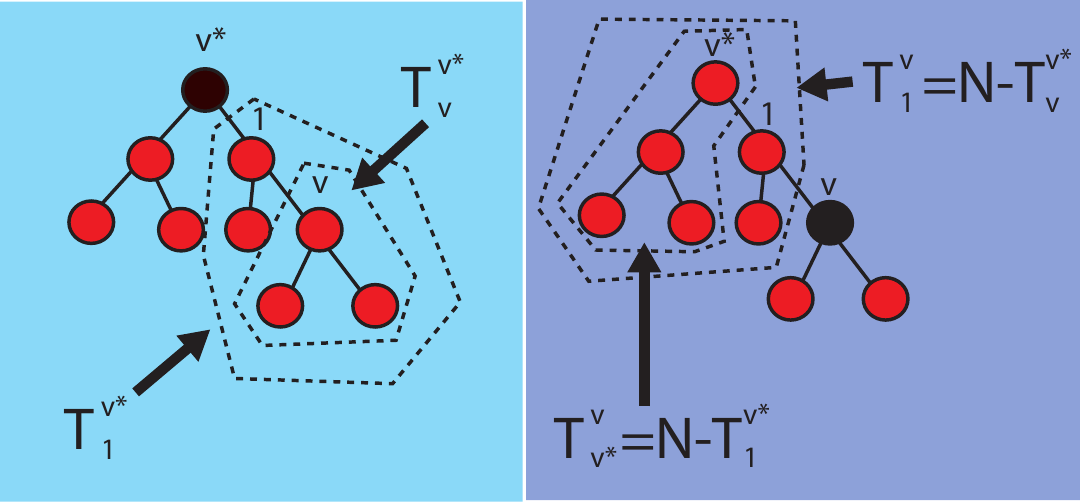}
		\caption{$T_i^j$ variables for source nodes 2 hops apart.}  
	\label{fig:tree-path}
\end{figure}
The following is an important characterization of the rumor center in
terms of the sizes of its local subtrees. As we shall see, this will 
play a crucial role in establishing our main results about the performance
of rumor centrality as an estimator for tree graphs. 
\begin{prop} \label{prop:rc}
Given an $N$ node tree, if node $v^*$ is the rumor center, 
then any subtree with $v^*$ as the source must have the following property:
\begin{equation}
	T^{v^*}_{v}\leq \frac{N}{2}.
\end{equation}
If there is a node $u$ such that for all $v\neq u$
\begin{equation}
	T^{u}_{v}\leq \frac{N}{2}
\end{equation}
then $u$ is a rumor center.  Furthermore, a tree can have at most 2 
rumor centers.
\end{prop}
\begin{proof}
We showed that for a tree with $N$ total nodes, for 
any neighboring nodes $u$ and $v$,
 \begin{equation}
      T_{u}^{v} = N-T^{u}_{v}.      
\end{equation}
For a node $v$ one hop from $v^*$, we find
\begin{equation*}
\displaystyle \frac{R(v,T)}{R(v^*,T)}=\frac{T^{v^*}_{v^*} T^{v^*}_{v}}{T_{v^*}^{v}T_{v}^{v}}
         =  \frac{T^{v^*}_{v}}{\left(N-T^{v^*}_{v}\right)}.
\end{equation*}
  When $v$ is two hops from $v^*$, all of the subtrees are the same except for those rooted at $v$, $v^*$, and the node in between, which we call node 1.  Figure \ref{fig:tree-path} shows an example.  In this case, we find
\begin{equation*}
	\frac{R(v,T)}{R(v^*,T)} = \frac{T^{v^*}_{v} T^{v^*}_{1}} {\left(N-T^{v^*}_{1}\right)\left(N-T^{v^*}_{v}\right)}.
\end{equation*}
Continuing this way, we find that in general, for any node $v$ in $T$,
\begin{equation}
	\frac{R(v,T)}{R(v^*,T)}=\displaystyle \prod_{i\in \mathcal P(v^*,v)}\frac{T^{v^*}_{i}} {\left(N-T^{v^*}_{i}\right)} \label{eq:path}
\end{equation}
where $\mathcal P(v^*,v)$ is the set of nodes in the path between $v^*$ and $v$, not including $v^*$.  

 Now imagine that $v^*$ is the rumor center.  Then we have
\begin{equation}
	\frac{R(v,T)}{R(v^*,T)}=\displaystyle \prod_{i\in \mathcal P(v^*,v)}\frac{T^{v^*}_{i}}. {\left(N-T^{v^*}_{i}\right)}\leq1
\end{equation}
For a node $v$ one hop from $v^*$, this gives us that
\begin{equation}
	T^{v^*}_{v}\leq \frac{N}{2}.\label{eq:rcn2}
\end{equation}
For any node $u$ in subtree $T^{v^*}_{v}$, we will have $T^{v^*}_{u}\leq T^{v^*}_{v}-1$.  Therefore, \eqref{eq:rcn2} will hold for any node $u\in T$.  This proves the first part of Proposition \ref{prop:rc}.  

Now assume that the node $v^*$ satisfies \eqref{eq:rcn2} for all $v\neq v^*$.  Then the ratios in \eqref{eq:path} will all be less than or equal to 1.  Thus, we find that 
\begin{equation}
	\frac{R(v,T)}{R(v^*,T)}=\displaystyle \prod_{i\in \mathcal P(v^*,v)}\frac{T^{v^*}_{i}} {\left(N-T^{v^*}_{i}\right)} \leq1.
\end{equation}
Thus, $v^*$ is the rumor center, as claimed in the second part of Proposition \ref{prop:rc}.  

Finally, assume that $v^*$ is a rumor center and that all of its subtrees satisfy $T_{v}^{v^*}<N/2$.  Then, any other node $v$ will have at least one subtree that is larger than $N/2$, so $v^*$ is the unique rumor center.  Now assume that $v^*$ has a neighbor $v$ such that $T_{v}^{v^*}=N/2$.  Then, $T_{v^*}^{v}=N/2$ also, and all other subtrees $T_{u}^{v}<N/2$, so $v$ is also a rumor center.  There can be at most 2 nodes in a tree with subtrees of size $N/2$, so a tree can have at most 2 rumor centers.
\end{proof}

\subsection{Rumor Centrality vs. Distance Centrality}

Here we shall compare rumor centrality with distance
centrality that has become popular in the literature 
as a graph based score function for various other 
applications. To start with, we recall the definition
of distance centrality. For a graph $G$, the distance 
centrality of node $v\in G$, $D(v,G)$, is defined as
\begin{equation}
	D(v,G)=\sum_{j\in G}d(v,j)
\end{equation}
where $d(v,j)$ is the shortest path distance from node $v$ to node $j$.  
The distance center of a graph is the node with the smallest distance 
centrality.  Intuitively, it is the node closest to all other nodes.  
On a tree, we will show the distance center is equivalent to the rumor center.  Therefore,
by establishing correctness of rumor centrality for tree graphs, one 
immediately finds that such is the case for distance centrality. 

We will prove the following proposition for the distance center of a tree.
\begin{prop} \label{prop:dc}
On an $N$ node tree, if $v_D$ is the distance center, then, for all $v\neq v_D$  
\begin{equation}
	T^{v_D}_{v}\leq \frac{N}{2}.
\end{equation}
Furthermore, if there is a unique rumor center on the tree, then it is 
equivalent to the distance center.
\end{prop}
\begin{proof}
Assume that node $v_D$ is the distance center of a tree $T$ which has $N$ nodes.  The distance centrality of $v_D$ is less than any other node.  We consider a node $v_\ell$ which is $\ell$ hops from $v_D$, and label a node on the path between $v_\ell$ and $v_D$ which is $h$ hops from $v_D$ by $v_h$.  Now, because we are dealing with a tree, we have the following important property.  For a node $j$ which is in subtree $T^{v_D}_{v_h}$ but not in subtree $T^{v_D}_{v_{h+1}}$, we have $d(v_\ell,j)=d(v_D,j)+d-2h$.  Using this, we find
\begin{align}
	   D(v_D,T)\leq& D(v_\ell,T) \nonumber\\
	   \sum_{j\in T}d(v_D,j) \leq& \sum_{v\in T}d(v_\ell,j) \nonumber\\
	   	   \leq& \sum_{j\in T}d(v_D,j) +\ell(N-T^{v_D}_{v_1})+ \nonumber\\
	   	    &(\ell-2)(T^{v_D}_{v_1}-T^{v_D}_{v_2})+...+(\ell-2\ell)(T^{v_D}_{v_\ell}) \nonumber\\
	   	\sum_{h=1}^\ell T^{v_D}_{v_h}  \leq& 	\sum_{h=1}^\ell(N-T^{v_D}_{v_h}).\label{eq:dc_sum}
\end{align}
If we consider a node $v_1$ adjacent to $v_D$, we find the same condition we had for the rumor center.  That is,
\begin{align}
	   	   T^{v_D}_{v_1}&\leq \frac{N}{2}. \label{eq:dc2}
\end{align}
For any node $u$ in subtree $T^{v_D}_{v_1}$, we will have $T^{v_D}_{u}\leq T^{v_D}_{v_1}-1$.   Therefore, \eqref{eq:dc2} will hold for any node $u\in T$.  This proves the first half of Proposition \ref{prop:dc}.  

If $v_D$ is a rumor center, then, it also satisfies \eqref{eq:dc2} as previously shown.  Thus, when unique, the rumor center is equivalent to the distance center on a tree.  This proves the second half of Proposition \ref{prop:dc}.
\end{proof} 

\begin{figure}
	\centering
		\includegraphics{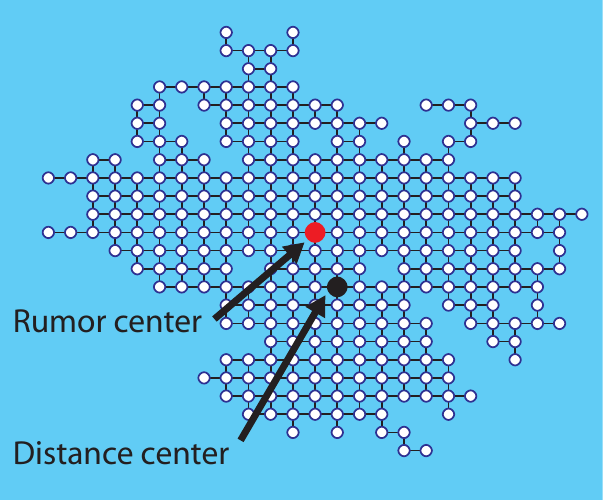}
		\caption{A network where the distance center does not equal the general graph rumor center.}  
	\label{fig:rc_vs_dc_picture}
\end{figure}

Now in contrast to trees, in a general non-tree network, the rumor center 
and distance center need not be equivalent. Specifically, we shall define
rumor centrality for a general graph to be the node with maximal value of rumor centrality on its own BFS tree.  Stated more precisely, the rumor center of a general graph is the node $\hat{v}$ 
with the following property (ties broken uniformly at random):
\begin{align}
	\hat{v} &\in \arg\max_{v\in G_N} R(v,T_{\bfs}(v)).
\end{align}
In a general graph, as can be seen in Figure \ref{fig:rc_vs_dc_picture}, this general graph rumor center is not always equivalent to the distance center as it was for trees.  We will see later that the general graph rumor center will be a better estimator of the rumor source than the distance center.  The intuition for this is that the distance center is evaluated using only the shortest paths in the graph, whereas the general graph rumor centrality utilizes more of the network structure for estimation of the source.
%

\subsection{Rumor Centrality and Linear Extensions of Posets}\label{ssec:poset}
The rumor graph on a network can be viewed as a partially ordered set, or {\em poset}, of nodes if we fix a source node as the root and consider the network to be directed, with edges pointing from the node that had the rumor to the node it infected.  These directed edges impose a partial order on the nodes.  We have referred to any permutation of the nodes which satisfies this partial order as a permitted permutation.  However, it is also known as a {\em linear extension} of the poset.  It is known that counting the number of linear extensions of a poset is in general a very hard problem (specifically, it falls in the complexity class \#P-complete \cite{ref:posetcomplex}).  However, on trees, counting linear extensions becomes computationally tractable.  To the best of our knowledge, 
the fastest known algorithm for counting linear extensions on a tree requires $O(N^2)$ 
computation \cite{ref:posetalg}.  In contrast, the message-passing algorithm we 
presented in Section \ref{ssec:algorithm} required only $O(N)$ computation.

\section{Main Results: Theory}\label{sec:mainres}
This section examines the behavior of the detection probability of the rumor source estimators for different graph structures.  We establish that the asymptotic detection probability has a phase-transition effect: for linear graphs it is 0, while for trees which grow faster than a line it is strictly greater than 0.  We will use different proof techniques to establish these results for trees with different rates of expansion.  

\subsection{Linear Graphs: No Detection}  We first consider the detection probability for a linear graph, which is a regular tree of degree 2.  We will establish the following result.
\begin{thm} \label{thm:line}
Define the event of correct rumor source detection after time $t$ on a linear graph as $\mathcal{C}_t$. Then the probability of correct detection of the ML rumor source estimator, $\mathbf{P}(\mathcal{C}_t)$, scales as
\begin{equation*}
	\mathbf{P}(\mathcal{C}_t)= O\left(\frac{1}{\sqrt{t}}\right).
\end{equation*}
\end{thm}
As can be seen, the linear graph detection probability scales as $t^{-1/2}$, which goes to 0 as $t$ goes to infinity.  The intuition for this result is that the estimator provides very little information because of the linear graph's trivial structure.

\subsection{Regular Expander Trees: Non-Trivial Detection} 
We next consider detection on a regular degree expander tree.  We assume each node has degree $d>2$.  For $d=2$, the tree is a line, and we have seen that the detection probability goes to 0 as the network grows in size.  For a regular tree with $d>2$ we obtain the following result.
\begin{thm} \label{thm:expander}
Define the event of correct rumor source detection by the ML 
rumor source estimator after time $t$ on a regular tree 
with degree $d>2$ as $\mathcal{C}_t$. Then there 
exists a constant $\alpha_d > 0$ for all $d > 2$ so that 
\begin{equation*}
0  ~<~	\alpha_d  ~\leq~ \liminf_{t}\mathbf{P}(\mathcal C_t) 
	~\leq~ \limsup_{t}\mathbf{P}(\mathcal C_t) 
	 ~\leq~ \frac{1}{2}. 
\end{equation*}
\end{thm}
Unlike linear graphs, when $d>2$ then there is enough 
`complexity' in the network that allows us to perform 
detection of the rumor source with strictly positive 
probability irrespective of $t$ (or size of the rumor
network). The above result also says that the detection
probability is always upper bounded by $1/2$ for any 
$d > 2$.

\subsection{Degree 3 Regular Expander Trees: Exact Detection Probability}

For regular trees of arbitrary degree $d > 2$, Theorem \ref{thm:expander}
states that the detection happens with strictly positive probability 
irrespective of the size of the network. However, we are unable to 
evaluate the exact asymptotic detection probability as $t \to\infty$. 
For $d = 3$, however we are able to obtain the exact value. 

\begin{thm} \label{thm:d3}
Define the event of correct rumor source detection under the
ML rumor source estimator after time $t$ on a regular expander 
tree with degree $d=3$ as $\mathcal{C}_t$. 
Then 
\begin{equation*}
	\lim_{t}\mathbf{P}(\mathcal C_t) = \frac{1}{4}.
\end{equation*}
\end{thm}

\subsection{Geometric Trees: Correct Detection}\label{ssec:geom}  

The above stated results cover the case of regular trees. We now 
consider the detection probability of our estimator in non-regular trees. 
As a candidate class of such trees, we consider trees that grow polynomially.
We shall call them geometric trees.  These non-regular trees are parameterized 
by constants $\alpha$, $b$, and $c$, with $0 < b \leq c$.  We fix a source 
node $v^*$ and consider each neighboring subtree of $v^*$. Let $d^*$ be the 
degree of $v^*$. Then there are $d^*$ subtrees of $v^*$, say $T_1,\dots,T_{d^*}$. 
Consider the $i$th such subtree $T_i$, $1\leq i\leq d^*$. Let $v$ 
be any node in $T_i$ and let $n^i(v,r)$ be the number of 
nodes in $T_i$ at distance exactly $r$ from the node $v$. Then 
we require that for all $1\leq i\leq d^*$ and $v \in T_i$
\begin{equation}
	b r^\alpha \leq n^i(v,r) \leq c r^{\alpha}. \label{eq:geom}
\end{equation}
The condition imposed by \eqref{eq:geom} states that each 
of the neighboring subtrees of the source should satisfy 
polynomial growth (with exponent $\alpha > 0$) and regularity 
properties. The parameter $\alpha > 0$ characterizes the 
growth of the subtrees and the ratio $c/b$ describes the 
regularity of the subtrees.  If $c/b\approx 1$ then the 
subtrees are somewhat regular, whereas if the ratio is much 
greater than 1, there is substantial heterogeneity in the subtrees.  

We note that unlike in regular trees, in a geometric tree 
the rumor centrality is not necessarily the ML estimator due
to the heterogeneity.  Nevertheless, we can use it as an
estimator. Indeed, as stated below we find that the rumor
centrality based estimator has an asymptotic detection probability of
$1$. That is, it is as good as the best possible estimator. 

\begin{thm}\label{thm:geom}
Consider a geometric tree as described above with 
parameters $\alpha > 0$, $0 < b \leq c$
that satisfy \eqref{eq:geom} for a node $v^*$ with degree $d_{v*} \geq 3$.  
Let the following condition be satisfied 
\begin{align*}
d_{v^*} & > \frac{c}{b} + 1.  
\end{align*}
Suppose the rumor starts spreading from node $v^*$ at time $0$ as per the SI
model. Let the event of correct rumor source detection with the rumor centrality based estimator after time $t$ 
in this scenario be denoted as $\mathcal{C}_t$. Then
\begin{equation*}
	\lim\inf_{t}\mathbf{P}(\mathcal{C}_t)=1.
\end{equation*}
\end{thm}
This theorem says that $\alpha=0$ and $\alpha>0$ serve as a threshold 
for non-trivial detection:  for $\alpha=0$, the graph is essentially a 
line graph, so we would expect the detection probability to go to $0$ 
as $t\to \infty$ based on Theorem \ref{thm:line}, but for $\alpha>0$ 
the detection probability converges to $1$ as $t\to\infty$.

\section{Simulation Results }\label{sec:simulations}
This section provides simulation results for our rumor source estimators on different network topologies.  These include synthetic topologies such as the popular scale-free and small-world networks, and also real topologies such as the Internet and the U.S. electric power grid.


\subsection{Tree Networks}  
\begin{figure}[t]
	\centering
		\includegraphics{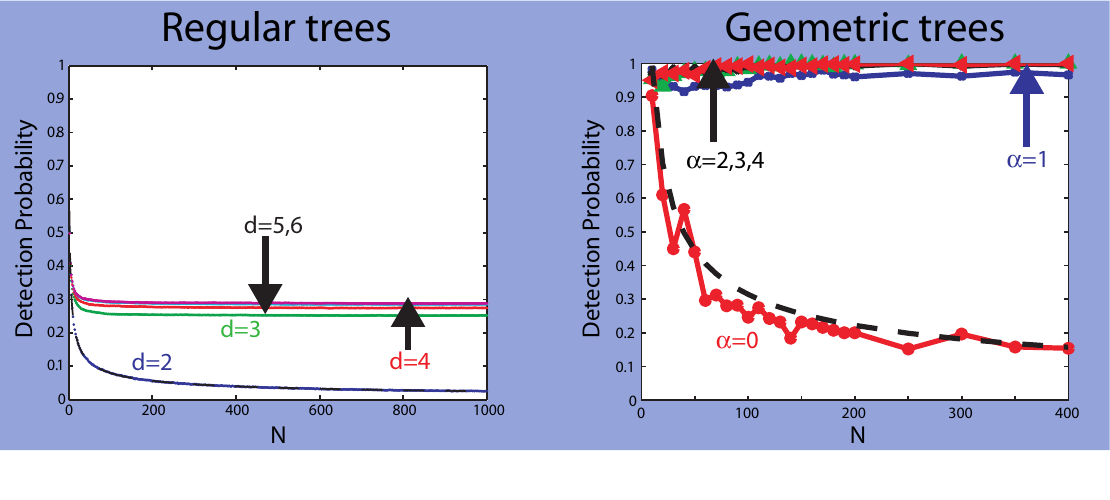}
	\caption{Rumor centrality detection probability for regular trees (left) and geometric trees (right) vs. number of nodes $N$.  The dotted lines are plots of $N^{-1/2}$.  
	}  
	\label{fig:linear-geom}
\end{figure}

The detection probability of rumor centrality versus network size for different trees is show in Figure \ref{fig:linear-geom}.  As can be seen, the detection probability decays as $N^{-1/2}$ as predicted in Theorem \ref{thm:line} for the graphs which grow like lines ($d=2$ and $\alpha=0$).  

For regular degree trees we see that the detection probability is less than $1/2$ and for $d\geq3$ it does not decay to $0$, as predicted by Theorem \ref{thm:expander}.  In fact, the detection probabilities appear to converge to asymptotic values.  This value is $1/4$ for $d=3$ as predicted by Theorem \ref{thm:d3}, and seems to increase by smaller amounts for $d=4,5,6$.

For geometric trees with $\alpha>0$, we see that the detection probability does not decay to 0 and is very close to 1 as predicted by Theorem \ref{thm:geom}.

\subsection{Synthetic Networks}  We performed simulations on synthetic small-world \cite{ref:sw} and scale-free \cite{ref:sf} networks.  These are two very popular models for networks and so we would like our rumor source estimator to perform well on these topologies.    For both topologies, the underlying graph contained 5000 nodes and in the simulations we let the rumor spread to 400 nodes.  

\begin{figure}[t]
	\centering
		\includegraphics{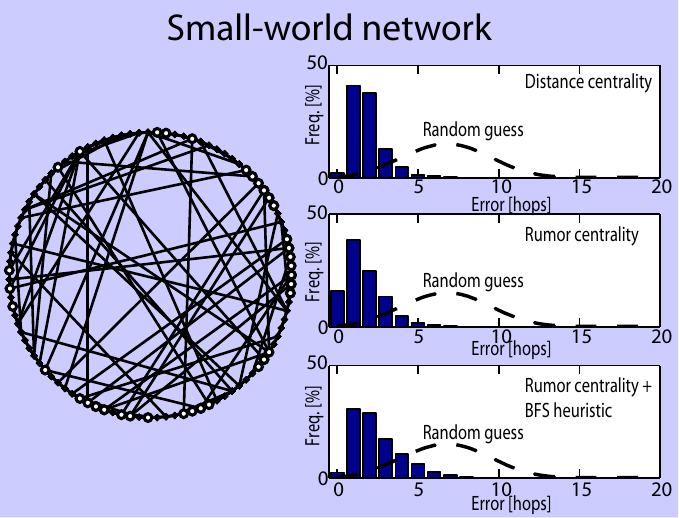}
		\caption{Histograms of the error for distance centrality, rumor centrality, and rumor centrality with BFS heuristic estimators on a 400 node rumor network on a small-world network.  The dotted line is a smooth fit of the histogram for randomly guessing the source in the rumor network.  An example of a rumor graph (infected nodes in white) is shown on the right.  
		}  
	\label{fig:sw}
\end{figure}

\begin{figure}[t]
	\centering
		\includegraphics{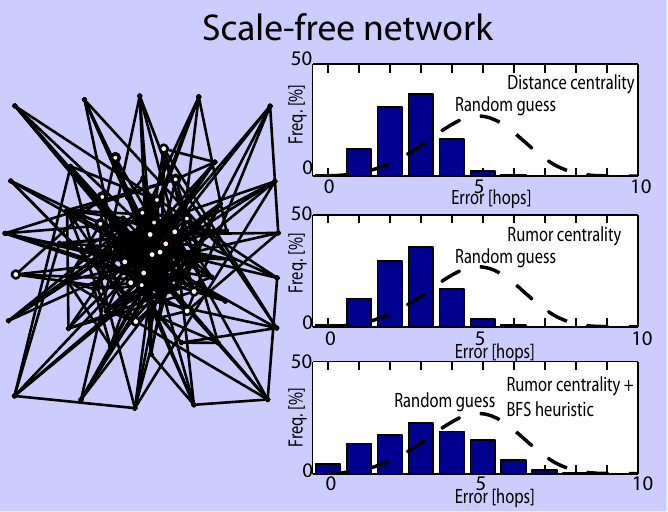}
		\caption{Histograms of the error for distance centrality, rumor centrality, and rumor centrality with BFS heuristic estimators on a 400 node rumor network on a scale-free network. The dotted line is a smooth fit of the histogram for randomly guessing the source in the rumor network.  An example of a rumor graph (infected nodes in white) is shown on the right.  
		}  
	\label{fig:sf}
\end{figure}

Figures \ref{fig:sw} and \ref{fig:sf} show an example of rumor spreading in a small-world and a scale-free network. The graphs show the rumor infected nodes in white.  Also shown are the histograms of the estimator error for three different estimators.  The estimators are distance centrality, rumor centrality on a BFS tree, and rumor centrality on a BFS tree with the BFS heuristic.  For comparison, we also show with a dotted line a smooth fit of the histogram for the error from randomly choosing the source from the 400 node rumor network.  As can be seen, for both networks, the histogram for the random guessing is shifted to the right of the estimator histograms.  Thus, the centrality based estimators are a substantial improvement over random guessing for both small-world and scale-free networks.

The distance centrality estimator performs very similarly to the rumor centrality estimator.  However, we see that on the small-world network, rumor centrality is better able to correctly find the source (0 error) than distance centrality (16\% correct detection versus 2\%).  For the scale-free network used here, the average ratio of edges to nodes in the 400 node rumor graphs is 1.5 and for the small-world network used here, the average ratio is 2.5.  For a tree, the ratio would be 1, so the small-world rumor graphs are less tree-like.    This may explain why rumor centrality does better than distance centrality at correctly identifying the source on the small-world network.  

The BFS heuristic leads to two visible effects.  First, as can be seen for the scale-free network, we have a larger correct detection probability.  Scale-free networks have power-law degree distributions, and thus contain many high degree {\em hubs}.  The BFS heuristic works well in these types of networks because it was precisely designed for networks with heterogeneous degree distributions.  

The second effect of the BFS heuristic is that larger errors become more likely.  For both networks, the histograms spread out to higher errors.  We see that for networks with less heterogeneous degree distributions, such as the small-world network, the BFS heuristic is actually degrading performance.  It may be that for more regular networks, the BFS heuristic amplifies the effect any slight degree heterogeneity and causes detection errors.

\subsection{Real Networks}  We performed simulations on an Internet autonomous system (AS) network \cite{ref:AS} and the U.S electric power grid network \cite{ref:sf}.  These are two important real networks so we would like our rumor source estimator to perform well on these topologies.  The AS network contained 32,434 nodes and the power grid network contained 4941 nodes.  In the simulations we let the rumor spread to 400 nodes.  

\begin{figure}[t]
	\centering
		\includegraphics{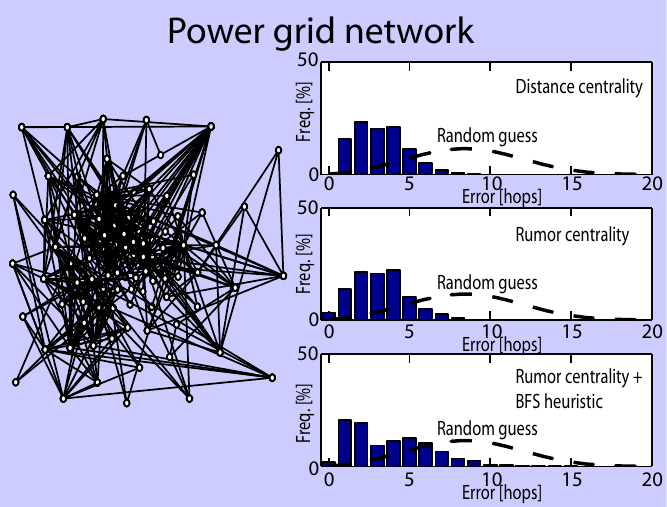}
		\caption{Histograms of the error for distance centrality, rumor centrality, and rumor centrality with BFS heuristic estimators on a 400 node rumor network on the U.S. electric power grid network. The dotted line is a smooth fit of the histogram for randomly guessing the source in the rumor network.  An example of a rumor graph (infected nodes in white) is shown on the right.  
		}  
	\label{fig:powergrid}
\end{figure}

\begin{figure}[t]
	\centering
		\includegraphics{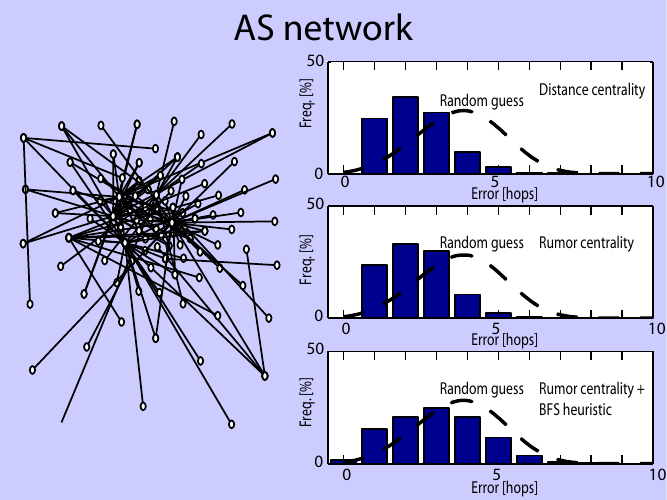}
		\caption{Histogram of the error for distance centrality, rumor centrality, and rumor centrality with BFS heuristic estimators on a 400 node rumor network on an Internet autonomous system (AS) network. The dotted line is a smooth fit of the histogram for randomly guessing the source in the rumor network.  An example of a rumor graph (infected nodes in white) is shown on the right.  
		}  
	\label{fig:AS}
\end{figure}

Figures \ref{fig:powergrid} and \ref{fig:AS} show an example of rumor spreading in both of these networks.  Also shown are the histograms of the estimator error for three different estimators.  The estimators are distance centrality, rumor centrality on a BFS tree, and rumor centrality on a BFS tree with the BFS heuristic.  For comparison, we also show with a dotted line a smooth fit of the histogram for the error from randomly choosing the source from the 400 node rumor network.  As with the synthetic networks, the histogram for the random guessing is shifted to the right of the estimator histograms.  Thus, on these real networks, the centrality based estimators are a substantial improvement over random guessing for both small-world and scale-free networks. 

We see that rumor centrality and distance centrality have similar performance, but for the power grid network, rumor centrality is better able to correctly find the source than distance centrality (3\% correct detection versus 0\%).  For the power grid network, the average ratio of edges to nodes in the 400 node rumor graphs is 4.2, and for the AS network the average ratio is 1.3.  Thus, the rumor graphs on the power grid network are less tree-like.  Similar to the small-world networks, this may explain why rumor centrality outperforms distance centrality on the power grid network.  

The BFS heuristic improves the correct detection probability for the AS network.  This is due to the fact that the AS network has many high degree hubs, similar to scale-free networks.  However, in the powergrid network, the BFS heuristic spreads out the histogram to higher errors.  Again, this may be due to the fact that the powergrid network does not have as much degree heterogeneity as the AS network, and the BFS heuristic is amplifying weak heterogeneities, similar to small-world networks.

\section{Proofs}\label{sec:proofs}

This section establishes the proofs of Theorems 
\ref{thm:line}-\ref{thm:geom}. All of them 
utilize rumor centrality as the estimator to 
obtain the desired conclusion. To study the property
of the rumor centrality, we shall utilize the
property of rumor center as established in 
Proposition \ref{prop:rc} crucially.

\subsection{Proof of Theorem \ref{thm:line}}

We consider the spread of the rumor in a line 
network starting from a source, say $v^*$. We 
shall establish that for any time $t > 0$, the 
probability of the rumor center being equal to
$v^*$ decays as $O(1/\sqrt{t})$. Since a line
is a regular graph and hence the rumor center is
the ML estimator, it follows that the detection
probability of any estimator decays as $O(1/\sqrt{t})$.
This is because in the absence of any prior 
information (or uniform prior) the ML estimator
minimizes the detection error (cf. see \cite{BT09}). 

Now rumor spreading in a line graph is equivalent to $2$ 
independent Poisson processes with rate $1$ beginning at 
the source and spreading in opposite directions. 
We refer to these processes as $N_1(t)$ and $N_2(t)$ with
$N_1(0) = N_2(0) = 0$. 

It follows from results of Section \ref{rc_prop} that 
the rumor center of a line is the center of the line: if the
line has an odd number of nodes then the rumor center is
uniquely defined, else there are two rumor centers.  Thus, 
we will correctly detect the source with probability $1$ 
if the two Poisson processes on each side of the source 
have exactly the same number of arrivals and with probability
$1/2$ if one of the Poisson processes is one less than the other.
Then, the probability of the event of correct detection at time $t$, 
$\mathcal{C}_t$ is given by 
\begin{align*}
 \mathbf{P}(\mathcal{C}_t)& = \mathbf P(N_1(t)=N_2(t)) \\
 & ~~ + \frac{1}{2} \big(\P(N_1(t)=N_2(t)+1) + \P(N_1(t)+1=N_2(t))\big)\\
                          &=  \sum_{k=0}^{\infty}\left(\mathbf e^{-t}\frac{t^k}{k!}\right)^2 + \left(\mathbf e^{-t}\frac{t^k}{k!}\right)^2 \frac{t}{k+1}. \\
\end{align*}
Let $a_k = e^{-2t} t^{2k}/(k!)^2$ and $b_k = a_k t/(k+1)$. Then
\begin{align*}
 \mathbf{P}(\mathcal{C}_t)& = \sum_{k=0}^\infty a_k + b_k. 
\end{align*}
We shall show that both $\sum_k a_k$ and $\sum_k b_k$ are
bounded as $O(1/\sqrt{t})$. This will conclude the proof 
of Theorem \ref{thm:line}. 

To that end, first we consider summation of $a_k$s. Let us
consider the ratios of the successive terms: 
\begin{equation*}
	\frac{a_{k}}{a_{k-1}}=\left(\frac{t}{k}\right)^2.
\end{equation*}
This ratio will be greater than $1$ as long as $k\leq t$ and beyond that
it will be less than $1$. Thus, $a_k$ is maximum for $k = \lfloor t \rfloor$. 
By Stirling's approximation, it follows that 
%
%
\begin{align*}
	\log(a_{\lfloor t\rfloor}) &= -2 \flt + 2 \flt \log(\flt)-2\log(\flt!) \\
	& =  -2 \flt + 2 \flt \log(\flt) - 2 \flt \log(\flt) \\
	& \qquad + 2\flt - \log(\flt) + \Theta(1) \\
	& = -\log(\flt) + \Theta(1). 
\end{align*}
Therefore, $a_{\flt} = \Theta\big(1/\flt\big)$. Given this, 
we shall bound all $a_k$s relative to $a_{\flt}$ to obtain
bound of $O(1/\sqrt{t})$ on $\sum_k a_k$. 

To that end, since $a_k$
is decreasing for $k \geq \flt$, we have that for any $k' \geq \flt$, 
\begin{equation*}
  \sum_{k=k'}^{k' + \sqrt{t}} a_k \leq \sqrt{t}a_{k'}. 
\end{equation*}
Similary, since for $k \leq \flt$, $a_k$ is increasing we have that for any
$k' \leq \flt -\sqrt{t}$, 
\begin{equation*}
  \sum_{k = k'}^{k' + \sqrt{t}} a_k \leq \sqrt{t}a_{k' + \sqrt{t}}. 
\end{equation*}
Given the above two inequalities, it will suffice to bound $a_{\flt + \ell \sqrt{t}}$
and $a_{\flt -\ell \sqrt{t}}$ for all $\ell \geq 0$ relate to $a_{\flt}$. For
this, consider the following. For $k \geq \flt$, 
\begin{align*}
		\frac{a_{k+ \sqrt{t}}}{a_{k}}& = \left(\frac{t^{\sqrt{t}}}{\prod_{j=1}^{\sqrt{t}} (k+j) }\right)^2\\
		       &=\prod_{j=1}^{\sqrt{t}} \Big(1+ \frac{j + (k-t)}{t}\Big)^{-2} \\
		       &\stackrel{(a)}{\leq} \prod_{j=1}^{\sqrt{t}} \Big(1+ \frac{j - 1}{t}\Big)^{-2} \\
		       & \stackrel{(b)}{\leq} \prod_{j=1}^{\sqrt{t}} e^{-\frac{(j-1)}{t}} \\
		       & = e^{-\frac{1}{2}+\frac{1}{2\sqrt{t}}} \\
		       & \leq e^{-\frac{1}{3}},  
\end{align*}
for $t \geq 36$. In above (a) follows from $k \geq \flt \geq t-1$,  
(b) follows from $1+x \geq e^{x/2}$ for $x \in [0,1]$ and $j\leq \sqrt{t}$. 
Similarly, for $k \geq \flt - \sqrt{t}$ 
\begin{align*}
		\frac{a_{k+ \sqrt{t}}}{a_{k}}& = \left(\frac{t^{\sqrt{t}}}{\prod_{j=1}^{\sqrt{t}} (k+j) }\right)^2\\
		       &=\prod_{j=1}^{\sqrt{t}} \Big(1+ \frac{j + (k-t)}{t}\Big)^{-2}\\
           & \stackrel{(c)}{\geq} \prod_{j=1}^{\sqrt{t}} \Big(1+ \frac{j}{t}\Big)^{-2}\\
           & \stackrel{(d)}{\geq} \prod_{j=1}^{\sqrt{t}} e^{-\frac{j}{t}} \\
           & \geq e^{-\frac{1}{2}},
\end{align*}
where (c) follows from $k \leq \flt \leq t$ and (b) from  $1+x \leq e^x$ for $x \geq 0$. 
From above, it follows that 
\begin{align*}
a_{\flt + \ell \sqrt{t}} & \leq e^{-\ell/3} a_{\flt}, \\
a_{\flt - \ell \sqrt{t}} & \leq e^{-\ell/2} a_{\flt}.
\end{align*}
Using all of the above discussion, it follows that for $t \geq 36$, 
\begin{align*}
   \sum_{k=1}^{\infty} a_k &\leq \sqrt{t} a_{\flt} \Big(\sum_{\ell=0}^{\infty} e^{-\ell/3} + e^{-\ell/2}\Big) \\
   & = O\Big(a_{\flt} \sqrt{t}\Big) \\
   & = O(1/\sqrt{t}). 
\end{align*}
In a very similar manner, it can be shown that $\sum_k b_k = O(1/\sqrt{t})$. 
Therefore, it follows that the probability of correct detection is bounded
as $O(1/\sqrt{t})$ in a line graph. This completes the proof of Theorem \ref{thm:line}.

\subsection{Proof of Theorem \ref{thm:expander}}

We wish to establish that given a $d \geq 3$ regular tree, the
probability of correct detection of the source using rumor
centrality, irrespective of $t$, is uniformly lower
bounded by a strictly positive constant, $\alpha_d > 0$ and upper bounded by $1/2$.

To find the lower bound on the detection probability, we shall utilize Proposition \ref{prop:rc} which
states the following. The source node has $d$ subtrees and
let $N_j(t) \geq 0, ~1\leq j\leq d$, be the number of 
rumor infected nodes in the $j$th subtree at time $t$. 
If all $N_j(t)$ are strictly less than half the total 
number of rumor infected nodes at time $t$, then the
source is the unique rumor center. Using this implication
of Proposition \ref{prop:rc}, we obtain the following 
bound on the event of correct detection $\mathcal C_t$:
\begin{align}
\left\{\omega\Big|\max_{1\leq i\leq d}N_j(t,\omega)\leq 
	             \frac{1}{2}\sum_{j=1}^{d}N_j(t,\omega)  \right\}
	     & \subseteq \mathcal C_t.        \label{eq:balance}
\end{align}
Therefore, by lower bounding the probability of the event on the left
in \eqref{eq:balance}, we shall obtain the desired lower bound $\P(\mathcal C_t)$.
Since $N_j(t)$ are independent and identically distributed due to
regularity of tree, we shall find the marginal distribution of $N_j(t)$. 
Now finding the precise `closed form' expression form the probability
mass function of $N_j(t)$ for all $d \geq 3$ seems challenging. Instead,
we shall obtain something almost close to that. 

To that end, consider $N_1(t)$. Let $T_n, ~n \geq 1,$ denote be the time 
between $n-1$st node getting infected and $n$th getting infected in $N_1(t)$. 
Since the source node is connected to the root of the first subtree
via the edge along with rumor starts spreading as per an exponential distribution
of rate $1$, it follows that $T_1$ has an exponential distribution of rate $1$. 
Once the first node gets infected in the subtree, the number of edges along
with the rumor can spread further is $d-1 = 1 + (d-2)$. More generally, every time
a new node gets infected, it brings in new $d-1$ edges and removes $1$ edge 
along with rumor can spread in the subtree. Now the spreading time along
all edges is independent and identically distributed as per an exponential 
distribution of rate $1$ and exponential random variables have the `memoryless'
property: if $X$ is an exponential random variable with rate $1$, then 
$\P(X > \zeta + \eta | X > \eta) = \P(X > \zeta)$ for any $\zeta, \eta \geq 0$. 
From above, we can conclude that $T_n$ equals the minimum of $1+(d-2)(n-1)$
independent exponential random variables of rate $1$. By the property of the 
exponential distribution, this equals an exponential random variable
of rate $1+(d-2)(n-1)$. Now let $S_n$ be the total time for $n$ nodes 
to get infected in the subtree, that is
\begin{equation}
	S_n = \sum_{i=1}^{n}T_i
\end{equation}
We state the following Lemma which states the precise density of $S_n$
and some of its useful properties. It's proof is presented later
in the section. 
\begin{lemma}\label{lemma:expander}
The density of $S_n$ for a degree $d$ regular tree, $f_{S_n}(t)$ is given by
\begin{align}
f_{S_n}(t) & =\begin{cases} e^{-t} & \mbox{~for}~ n =1 \\ \Big(\prod_{i=1}^{n-1}\Big(1+\frac{1}{a i}\Big)\Big) e^{-t}(1-e^{-at})^{n-1} & \mbox{~for}~n \geq 2,
\end{cases} 
\end{align}
where $a=d-2$.  Further, let $\tau_{n}=\frac{1}{a} \big(\log(n) + \log(\frac{3}{4a})\big)$ 
and $t_n=\frac{1}{a} \log((n-1)a+1)$. Then 
\begin{itemize}
  \item[0.] $\tau_n \leq t_{n-1} \leq t_n$ for all $n \geq 2$. 
	\item[1.] For all $n \geq 1$ and $t \in (0, t_n)$, 
	\[ \frac{df_{S_n}(t)}{dt} > 0. \]
	
	\item[2.] There exists finite constants $B_a, C_a > 0$ so that 
	\[ \liminf_{n} f_{S_n}(\tau_n)\geq B_a, \mbox{~and~} \limsup_{n} f_{S_n}(t_n)\leq C_a.\] 
	
	\item[3.] There exists $\gamma \in (0,1)$ so that for all $t \in (0,t_n)$ 
	\[ \limsup_{n} \frac{f_{S_{(d-1)n}}(t)}{f_{S_n}(t)}\leq(1-\gamma). \]
	\end{itemize}
\end{lemma}

Next we use Lemma \ref{lemma:expander} to obtain a lower bound on
$\P(\mathcal C_t)$. For this, define $\mathcal D_n(t)$ as the event under
which all the $d$ subtrees have between $n$ and $(d-1)n$ infected nodes at 
time $t$.  That is, 
\begin{equation}
	\mathcal D_n(t) = \bigcap_{j=1}^{d} \left\{ n ~\leq~ N_j(t)~\leq~ (d-1)n\right\}, \mbox{~for~} n\geq 0.
\end{equation}
Under $\mathcal D_n(t)$ for any $n \geq 0$, it follows that $N_j(t) \leq \frac{1}{2} \sum_{j'=1}^d N_{j'}(t)$
for all $1\leq j \leq d$. That is, event $\mathcal C_t$ holds. Therefore, 
\begin{equation}\label{eq.lb.1}
     \mathbf{P}(\mathcal C_{t}) \geq \sup_{n\geq 0} \mathbf{P}(\mathcal D_n(t)).
\end{equation}
Therefore, to uniformly lower bound $\P(\mathcal C_t)$ and establish 
Theorem \ref{thm:expander}, it is sufficient to find $n(t)$ so that
$\P(\mathcal D_{n(t)}(t)) \geq \alpha_d > 0$ for all $t$ large enough. 
To that end, we shall first study how to bound $\mathbf{P}(\mathcal D_n(t))$.  
Under $\mathcal D_n(t)$, we have $ n \leq N_j(t) \leq (d-1)n$ for $1\leq j\leq d$.
Now consider $N_1(t)$. For $n\leq N_1(t) \leq (d-1)n$, it must be that 
the $n$th node in it must have got infected before $t$ and the 
$(d-1)n+1$st node must have got infected after $t$. Using this
along with the independent and identical distribution of $N_j(t)$ 
for $1\leq j\leq d$, we obtain
\begin{align}
\mathbf{P}\big(\mathcal D_{n}(t)\big)& = \P\Big(\bigcap_{j=1}^d n\leq N_j(t) \leq (d-1)n \Big) \nonumber \\
& = \P\Big(n\leq N_1(t) \leq (d-1)n\Big)^d \nonumber \\
& = \P\Big(S_n \leq t < S_{(d-1)n}\Big)^d \nonumber\\
 & =  \Big(\P\big(S_n \leq t\big) - \P\big(S_{(d-1)n} \leq t\big)\Big)^d\nonumber\\
 & = \left(\int_{0}^{t} f_{S_n}(\tau)-f_{S_{(d-1)n}}(\tau) ~d\tau\right)^d\label{eq:integral}
\end{align}
We shall use Lemma \ref{lemma:expander} to uniformly lower bound \eqref{eq:integral} 
by a strictly positive constant $\alpha_d > 0$ for all $t$ large enough. To that end, 
consider a large enough $t$ and hence $n \geq 2$ large enough so that 
using Lemma \ref{lemma:expander}, we have for any given 
small enough $\delta \in (0,1)$: (a) $\tau_n \leq t_{n-1} \leq t \leq t_n$, 
(b) $f_{S_n}(\tau_n)\geq B_a(1-\delta)$, (c) $f_{S_n}(t_n)\leq C_a(1+\delta)$ and (d) 
$f_{S_{(d-1)n}}(t)\leq \big(1-\gamma(1-\delta)\big) f_{S_n}(t)$ for all 
$t\in (0,t_n)$. Then using $\gamma' = \gamma (1- \delta)$, 
\begin{align*}
   & \int_{0}^{t}(f_{S_n}(\tau)-f_{S_{(d-1)n}}(\tau)) ~d\tau  \\
   & \quad \geq \int_{0}^{t}\left(f_{S_n}(\tau)-\big(1-\gamma'\big)f_{S_n}(\tau)\right) ~d\tau\\
   & \quad \geq \gamma' \int_{0}^t f_{S_n}(\tau) ~d\tau \\
   & \quad \geq \gamma'\Big(\int_{\tau_n}^{t_n} f_{S_n}(\tau)~d\tau - \int_{t_{n-1}}^{t_n} f_{S_n}(\tau) ~d\tau\Big)\\
   & \quad \geq \gamma' \Big(f_{S_n}(\tau_n)(t_n-\tau_n)- f_{S_n}(t_n)(t_n-t_{n-1})\Big)\\
   & \quad \geq \gamma' (1-\delta) B_a \frac{1}{a} \log\Big(\frac{4}{3}\Big) \\
   & \quad \qquad ~-~
              \gamma C_a (1+\delta) \frac{1}{a}\log\Big(\frac{(n-1)a + 1}{(n-2)a + 1}\Big).
\end{align*}
Above we have used the fact that for $n \geq 1$, 
$t_n - \tau_n \geq \frac{1}{a} \log \frac{4a}{3}$ and $a = d-2 \geq 1$. 
As $t \to \infty$, the corresponding $n$ so that $\tau_n \leq t_{n-1}\leq t \leq t_n$
increases to $\infty$ as well. Since the choice of $\delta \in (0,1)$ is arbitrary, from
above along with \eqref{eq.lb.1} and \eqref{eq:integral}, it follows that 
\begin{align*}
\liminf_n\mathbf{P}(\mathcal C_{t}) &\geq  \Big(\frac{1}{a} \gamma B_a \log \Big(\frac{4}{3}\Big)\Big)^d \\
                                    & \stackrel{\triangle}{=} \alpha_d ~>~0.
 \end{align*}
 
\begin{figure}[t]
	\centering
		\includegraphics{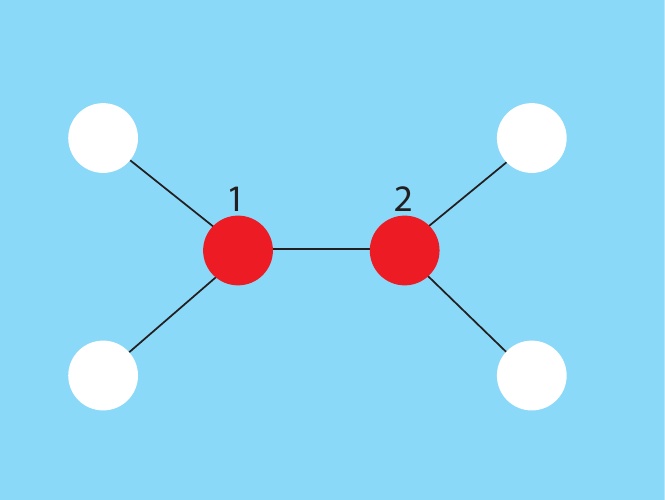}
		\caption{Symmetric 2 node rumor network in a regular tree.}  
	\label{fig:boundary}
\end{figure}

Next, we consider the upper bound of $1/2$.  This bound can be obtained by using symmetry arguments.  First imagine that the rumor has spread to two nodes.  First, because of the memoryless property of the spreading times, the spreading process essentially resets after the second node is infected, so we can treat these two nodes as just a single, enlarged rumor source.  Second, because of the regularity of the tree, the rumor boundary is symmetric about these two nodes, as shown in Figure \ref{fig:boundary}.  Therefore, within this enlarged rumor source, the estimator will not be able to distinguish between these two nodes due to symmetry.  For example, in Figure \ref{fig:boundary}, the estimator will select node 1 or node 2 with equal probability.  In the best scenario, the estimator will detect this enlarged rumor source exactly with probability 1. This happens for example, when the rumor network only has 2 nodes as in Figure \ref{fig:boundary}.  Then due to symmetry, the probability of correctly detecting the source is $1/2$ since each node in the enlarged rumor source is chosen with equal probability.  The probability of the estimator detecting the enlarged source is no greater than 1 ever, so the correct detection probability can never be greater than $1/2$.
 
This completes the proof of Theorem \ref{thm:expander}.


\medskip 

\begin{proof}[Proof of Lemma \ref{lemma:expander}]
We derive the density by induction.  For $n=1$, we trivially
have 
\begin{equation}
	f_{S_1}(t)=e^{-t}.
\end{equation}
Now, inductively assume that $f_{S_n}$ has the
form as claimed in Lemma \ref{lemma:expander}. That
is, 
\[ 
f_{S_n}(\tau) = C(n) e^{-\tau} \big(1-e^{-a\tau}\big)^{n-1}, ~\mbox{for}~\tau \geq 0,
\]
with $C(n) = \prod_{i=1}^{n-1} \Big(1+\frac{1}{a i}\Big)$.  
Now $S_{n+1} = S_n + T_{n+1}$; $T_{n+1}$ is independent of $S_n$
and has exponential distribution with rate $1 + (d-2) n = 1 + an$.
Therefore,  
\begin{align*}
  f_{S_{n+1}}(t)& = \int_{0}^t f_{S_n}(\tau) f_{T_{n+1}}(t-\tau)~d\tau \\
                & = C(n) (1+an) \int_{0}^{t}e^{-\tau}(1-e^{-a\tau})^{n-1} e^{-(1+an)(t-\tau)}~d\tau\\
                & = C(n+1)an e^{-(1+an)t} \Big(\int_{0}^{t}e^{an\tau}
                   (1-e^{-a\tau})^{n-1}~d\tau\Big).                       
\end{align*}
Expanding $(1-e^{-a\tau})^{n-1} = \sum_i {n-1\choose i}(-1)^i e^{-ai \tau}$ and integrating,
we obtain
\begin{align*}
 \int_{0}^{t} e^{an\tau}
                   (1-e^{-a\tau})^{n-1}d\tau &=
                   \sum_{i=0}^{n-1}{n-1\choose i}(-1)^i\frac{e^{a (n-i) t}-1}{a(n-i)} \\
                   & = \frac{1}{an} \sum_{i=0}^{n-1} {n \choose i} (-1)^i \big(e^{a (n-i) t}-1\big). 
\end{align*}
From this and above, we obtain
{\small
\begin{align*}
 & f_{S_{n+1}}(t) \\
  & = C(n+1) e^{-(1+an)t}\sum_{i=0}^{n-1}{n\choose i}(-1)^i\big(e^{a(n-i)t}-1\big)\\
    & =  C(n+1)e^{-t} \sum_{i=0}^{n-1}{n\choose i}(-1)^i \big(e^{-a i t}- e^{-a n t}\big)\\
    & = C(n+1) e^{-t} \Big(\Big(\sum_{i=0}^n {n \choose i} \big(-e^{-at}\big)^i\Big) -  \big(-e^{-at}\big)^n\Big) \\
    & \quad - C(n+1) e^{-t} e^{-ant} \Big(\Big(\sum_{i=0}^n {n\choose i} (-1)^i \Big) - (-1)^n\Big) \\
    & = C(n+1) e^{-t} \Big( (1-e^{-at})^n - (-e^{-at})^n + (-e^{-at})^n \Big) \\
    & = C(n+1) e^{-t} (1-e^{-at})^n. 
\end{align*}
}
This is precisely the form claimed by Lemma \ref{lemma:expander}. This completes
the induction step and establishes the form of density of $S_n$ as claimed. Next
we establish four properties claimed in Lemma \ref{lemma:expander}. 


\medskip
\noindent{\em Property 0.} Let $n \geq 8$. Clearly, $t_{n-1} \leq t_n$ as $a = d-2 \geq 1$
for $d \geq 3$. Now for $n \geq 8$, we have $an/4 \geq 2a-1$. Therefore, $(n-2)a + 1 \geq 3an/4$.
Therefore, it follows that $\tau_n \leq t_{n-1}$. 

\medskip
\noindent{\em Property 1.} Consider any $n \geq 1$ and $t \in (0,t_n)$ with $t_n = \frac{1}{a} \log ((n-1)a+1)$. 
Now 
\begin{align*}
& \quad ~~~~t < t_n \\
& \Rightarrow ~~t < \frac{1}{a} \log ((n-1) a + 1) \\
& \Rightarrow ~~0 < e^{-at} ((n-1) a + 1) - 1 \\
& \Rightarrow ~~0 < \frac{df_{S_n}(t)}{dt}.  
\end{align*}
In above, we have used the form of $f_{S_n}$ established earlier. 

\medskip 
\noindent{\em Property 2.} Our interest is in obtaining 
uniform upper bound on $f_{S_n}(t_n)$ and uniform lower
bound on $f_{S_n}(\tau_n)$. To that end, let us start with
the following standard inequality. 
\begin{align*}
1+ \frac{1}{a} \log n & \geq \frac{1}{a}\sum_{i=1}^{n-1}\frac{1}{i} \\
                      &	\geq \log\left(\prod_{i=1}^{n-1}\left(1+\frac{1}{ai} \right)\right) \\
                      & \geq \frac{1}{a} \sum_{i=1}^{n-1}\left(\frac{1}{i}-\frac{1}{i^2}\right)\\	 
                      & \geq \frac{1}{a}\log(n-1) - \frac{\zeta(2)}{a}         
\end{align*}
where $\zeta(2) = \sum_{i=1}^\infty i^{-2} = \pi^2/6$. Recall that 
$C(n) = \prod_{i=1}^{n-1} \big(1+\frac{1}{a i}\big)$. Therefore, from above we have
\begin{equation}\label{eq:Cn}
e n^{\frac{1}{a}} ~\geq~ C(n) ~\geq~(n-1)^{\frac{1}{a}} e^{-\frac{\zeta(2)}{a}}. 
\end{equation}
Recalling $\tau_n = \frac{1}{a} \log \frac{3n}{4a}$ and from above, 
$f_{S_n}(\tau_n)$ can be lower bounded as 
\begin{align*}
	   f_{S_n}(\tau_n) & = C(n) e^{-\tau_n} \big(1-e^{-a\tau_n}\big)^{n-1} \\
	                   & = C(n) \Big(\frac{4a}{3n}\Big)^{\frac{1}{a}} \Big(1-\frac{4a}{3n}\Big)^{n-1} \\
	                   & \geq  e^{-\frac{\zeta(2)}{a}} \Big(\frac{4a (n-1)}{3n}\Big)^{\frac{1}{a}} \Big(1-\frac{4a}{3n}\Big)^{n-1}. 
\end{align*}
Now the term $e^{-\frac{\zeta(2)}{a}}$ is strictly positive constant; 
$(\frac{n-1}{n})^{\frac{1}{a}} \to 1$ as $n\to\infty$; and 
$\big(1-\frac{4a}{3n}\big)^{n-1} \to e^{-\frac{4a}{3}}$ as $n\to\infty$. 
Therefore, it follows that 
\begin{align*}
\liminf_{n\to\infty} f_{S_n}(\tau_n) & \geq e^{-\frac{\zeta(2)}{a}}\Big(\frac{4a}{3}\Big)^{\frac{1}{a}} e^{-\frac{4a}{3}} \\
& \stackrel{\triangle}{=} B_a ~>~0. 
\end{align*}

Similarly, for $t_n = \frac{1}{a} \log ((n-1)a + 1)$, 
\begin{align*}
	   f_{S_n}(t_n) & = C(n) e^{-t_n} \big(1-e^{-at_n}\big)^{n-1} \\
	                & = C(n) \Big(\frac{1}{(n-1)a+1}\Big)^{\frac{1}{a}} \Big(1-\frac{1}{(n-1)a+1}\Big)^{n-1} \\
	                & \leq e \Big(\frac{n}{(n-1)a+1}\Big)^{\frac{1}{a}} \Big(1-\frac{1}{(n-1)a+1}\Big)^{n-1} \\
	                & \stackrel{n\to\infty}{\longrightarrow} e a^{-\frac{1}{a}} e^{-\frac{1}{a}} \\
	                & \stackrel{\triangle}{=} C_a ~< \infty. 
\end{align*}
Thus 
\begin{equation}
	\limsup_n f_{S_n}(t_n) ~\leq~ C_a ~<~ \infty.
\end{equation}

\medskip 
\noindent{\em Property 3.} To establish this property, recall that for $m \geq 2$ 
\begin{align*}
\log f_{S_m}(t) & = \sum_{i=1}^{m-1} \log \Big(1+\frac{1}{a i}\Big) - t \\
&\qquad + (m-1) \log \Big(1-e^{-at}\Big). 
\end{align*}
Therefore (using $a = d-2$)
\begin{align*}
	& \log\left(\frac{f_{S_{(d-1)n}}(t)}{f_{S_n}(t)} \right) \\
	& ~~= \sum_{i=n}^{(a+1)n-1} \log\left(1+\frac{1}{a i} \right) + a n \log(1-e^{-at})\\
	      &~~\leq \frac{1}{a}\Big(\sum_{i=n}^{(a+1)n-1}\frac{1}{i}\Big) + a n \log(1-e^{-a t})\\
	      &~~\leq \frac{1}{a}\Big(\log\Big(\frac{(a+1)n}{n}\Big)\Big) + a n \log(1-e^{-a t}).\\
	      &~~= \frac{1}{a} \log (a+1) + a n \log(1-e^{-a t}).
\end{align*}
For $t \in (0,t_n)$, $e^{-a t} \geq e^{-a t_n}$ and hence 
\[ e^{-a t} \geq e^{-a t_n} ~=~\frac{1}{(n-1) a + 1} ~\geq~\frac{1}{na},  \]
since $a \geq 1$. Therefore, for $t \in (0,t_n)$ we have 
\begin{align*}
	& \log\left(\frac{f_{S_{(d-1)n}}(t)}{f_{S_n}(t)} \right) \\
	&~~\leq \frac{1}{a} \log (a+1) + a n \log \Big(1-\frac{1}{an}\Big) \\
	&~~\stackrel{(x)}{\leq} \frac{1}{a} \log (a + 1) + a n \Big(-\frac{1}{a n} + \frac{1}{2 a^2 n^2}\Big) \\
	&~~\leq \frac{1}{a} \log (a + 1) - 1 + \frac{1}{2 a n},
\end{align*}
where (x) follows from $\log (1-x) \leq -x + x^2/2$ for all $x \in (0,1)$ and $an \geq 1$. 
Now $\frac{1}{2an} \to 0$ as $n\to\infty$; $\log (a+1)-a$ is a decreasing
function in $a \geq 0$ and for $a = 1$, it is $\log 2 - 1 \leq 1/6$. Therefore, it 
follows that there exists a $\gamma \in (0,1)$ so that for any $a \geq 1$, 
\begin{align*}
	\limsup_{n\to\infty} \log\left(\frac{f_{S_{(d-1)n}}(t)}{f_{S_n}(t)} \right) 
	& \leq \frac{1}{a} \log (a+1) - 1  \\
  & \leq \log (1-\gamma). 
\end{align*}
This completes the proof of Lemma \ref{lemma:expander}. 
\end{proof}

\subsection{Proof of Theorem \ref{thm:d3}}

We are interested in degree $d = 3$ regular tree. 
By Proposition \ref{prop:rc}, the event of correct detection 
at time $t$, $\mathcal C_t$ is such that 
\begin{align}\label{eq:d.3.lb}
	\left\{\omega\Big|\max_{j\in {1,2,3}}N_j(t,\omega)\leq 
	             \frac{1}{2}\sum_{j=1}^{3}N_j(t,\omega)  \right\} & \subseteq \mathcal C_t 
\end{align}
and 
\begin{align*}\label{eq:d.3.ub}
	\mathcal C_t \subseteq \left\{\omega\Big|\max_{j\in {1,2,3}}N_j(t,\omega)\leq 
	             \frac{1}{2}\Big(\sum_{j=1}^{3}N_j(t,\omega)\Big) + 1 \right\}.  
\end{align*}
If we knew the exact form for the distribution of the number of 
arrivals at time $t$, $N_j(t)$ for $1\leq j\leq 3$, 
then we could bound the probability of $\mathcal C_t$
explicitly. For regular trees with degree $3$, this is indeed 
possible.  To that end, by Lemma \ref{lemma:expander} we find 
that the distribution of the time for $n$ nodes to get rumor 
infected in the $j$th subtree, for $1\leq j\leq 3$ is given by 
\begin{align*}
	f_{S_n}(t)
	          &= ne^{-t}(1-e^{-t})^{n-1}, ~~\mbox{for}~n\geq 1.
\end{align*}
Now in order for there to be exactly $n$ rumor infected nodes 
in the $j$th subtree by time $t$, the $n$th node must get infected 
before $t$ and $n+1$st node must get infected after $t$.  Therefore,
the distribution of $N_j(t)$, for $1\leq j\leq 3$, is 
\begin{align*}
	\mathbf P\big(N_j(t)=n \big) & = \P\big(S_n \leq t\big)-\P\big(S_{n+1}\leq t\big)\\
	          & = \int_{0}^{t} n e^{-\tau}(1-e^{-\tau})^{n-1}d\tau \\ 			 
	          &\qquad -\int_{0}^{t}(n+1)e^{-\tau}(1-e^{-\tau})^{n} d\tau\\
	          & = e^{-t}(1-e^{-t})^{n},
\end{align*}
for $n \geq 1$. Indeed, 
\begin{align*}
\P\big(N_j(t) = 0\big) & = \P(S_1 > t) ~=~e^{-t}.
\end{align*}
If we denote $e^{-t}$ by $p$ then the above becomes 
\begin{align}
	\P\big(N_j(t)=n\big)&= p(1-p)^{n}, ~~\mbox{for}~~n\geq 0.
\end{align} 
That is, $N_i(t)$ has a geometric distribution with parameter $p = e^{-t}$. Next
we evaluate lower bound on $\P(\mathcal C_t)$ using \eqref{eq:d.3.lb}. To that end
define $\mathcal H_t^n$ as 
\[
\mathcal H_t^n = \left\{\omega\,\Big{|}\,\sum_{j=1}^{3}N_j(t,\omega)=n \bigcap \max_{j=1,2,3}N_j(t,\omega)\leq \frac{n}{2} \right\}.
\]
Then from \eqref{eq:d.3.lb} 
\begin{equation}
\bigcup_{n=0}^{\infty}\mathcal H_t^n \subseteq \mathcal C_t.
\end{equation}
Therefore, using the fact that the spreading on subtrees happens independently, 
we obtain
\begin{align}
	\mathbf P(\mathcal C_t)& \geq \sum_{n=0}^{\infty}\sum_{n_1,n_2,n_3\in \mathcal H_t^n}
	            \prod_{j=1}^{3}P(N_j(t)=n_j)\nonumber\\
	            & = \sum_{n=0}^{\infty}\sum_{n_1,n_2,n_3\in \mathcal H_t^n}p^{3} (1-p)^{n_1+n_2+n_3}\nonumber\\
	            &=  p^3\sum_{n=0}^{\infty}(1-p)^n\sum_{n_1,n_2,n_3\in \mathcal H_t^n}1.\label{eq:cnt}
\end{align} 
The sum over the $n_j$'s require us to count the number of states in $\mathcal H_t^n$. It
follows that 
\begin{align*}
|\mathcal H_t^n| & \geq 3! \times |\{(n_1,n_2,n_3): ~n_3 < n_2 < n_1 \leq n/2\}| \\
                & = 6 |\{(n_1,n_2,n_3): ~n_3 < n_2 < n_1 \leq n/2\}|.
\end{align*}
Now when $n_3 < n_2 < n_1 \leq n/2$, it must be that $n/3 \leq n_1 \leq n/2$; 
for a given such $n_1$, $(n-n_1)/2 \leq n_2 < n_1$ and $n_3 = n-n_1-n_2$. Using
these relations, it follows that the number of such $(n_1,n_2,n_3)$ triples are
$\frac{n^2}{48} + O(n)$. Therefore, $|\mathcal H_t^n| $ is at least $\frac{n^2}{8} + O(n)$.
Using this, we obtain  
\begin{align*}
\P\big(\mathcal C_t\big) & \geq p^3 \Big(\sum_{n=0}^{\infty}(1-p)^n \Big(\frac{n^2}{8} + O(n)\Big)\Big)\\
	         &= \frac{1}{8} p^3(1-p)^2 \Big(\sum_{n=0}^\infty (n+2)(n+1)(1-p)^{n}\Big) \\
	         & \qquad + p^3(1-p) O\Big(\sum_{n=0}^\infty n (1-p)^{n-1}\Big) \\
	         & \quad \qquad + p^3 O\Big(\sum_{n=0}^\infty (1-p)^n\Big) \\
	         & = \frac{p^3(1-p)^2}{8} \frac{2}{p^3} +  p^3 (1-p) O\big(p^{-2}\big) + p^3 O\big(p^{-1}\big) \\
	         & = \frac{(1-p)^2}{4} + O\big(p (1-p) + p^2\big). 
\end{align*} 
Now since $p = e^{-t}$, as $t\to\infty$, $p \to 0$. Therefore, we obtain that
\begin{align*}
\liminf_{t\to\infty} \mathbf P(\mathcal C_t)&\geq \frac{1}{4}. 
\end{align*}
In a very similar manner, using \eqref{eq:d.3.ub} it follows that
\begin{align*}
\P\big(\mathcal C_t\big) & \leq p^3 \Big(\sum_{n=0}^{\infty}(1-p)^n \Big(\frac{n^2}{8} + O(n)\Big)\Big) \\
                         & = \frac{(1-p)^2}{4} + O\big(p (1-p) + p^2\big),
\end{align*}
where $p = e^{-t}$. Therefore, 
\begin{align*}
\limsup_{t\to\infty} \mathbf P(\mathcal C_t)&\leq \frac{1}{4}. 
\end{align*}
This concludes the proof of Theorem \ref{thm:d3}. 

\subsection{Proof of Theorem \ref{thm:geom}}

The proof of Theorem \ref{thm:geom}, as before, uses
the characterization of rumor center provided by 
Proposition \ref{prop:rc}. That is, we wish to
show that for all $t$ large enough, the probability
of the event that the size of the $d^*$ rumor infected sub-trees 
of the source $v^*$ are 
essentially `balanced' (cf. \eqref{eq:balance}) 
with high enough probability. 
To establish this, we shall use coarse estimations
on the size of each of these sub-trees using the standard
concentration property of the Poisson process along with
geometric growth. This will be unlike the proof for 
regular trees where we had to necessarily delve into 
very fine detailed probabilistic estimates of the 
size of the sub-trees to establish the result. This 
relatively easier proof for geometric trees (despite 
heterogeneity) brings out the fact that it is 
fundamentally much more difficult to analyze expanding
trees than geometric structure as expanding trees do 
not yield to generic concentration based estimations
as they necessarily have very high variances. 

To that end, we shall start by obtaining sharp estimations on
the size of each of the rumor infected $d^*$ sub-trees of
$v^*$ for any given time $t$. Now initially, at time $0$
the source node $v^*$ has the rumor. It starts spreading along 
its $d^*$ children (neighbors). Let $N_i(t)$ denote the size
of the rumor infected subtree, denoted by $G_i(t)$, 
rooted at the $i$th child (or neighbor) of node $v^*$. 
Initially, $N_i(0) = 0$. The $N_i(\cdot)$ is a Poisson 
process with time-varying rate: the rate at time $t$ 
depends on the `boundary' of the tree as discussed 
earlier. Due to the balanced and geometric growth 
conditions assumed in Theorem \ref{thm:geom}, the following
will be satisfied: for small enough $\epsilon > 0$
(a) every node within a distance $t\left(1-\epsilon\right)$ 
of $v^*$ is in one of the $G_i(t)$, and (b) no node beyond distance 
$t\left(1+\epsilon\right)$ of $v^*$ is in any of the $G_i(t)$. 
Such a tight characterization of the `shape' of $G_i(t)$ along with
the polynomial growth will provide sharp enough bound on $N_i(t)$
that will result in establishing Theorem \ref{thm:geom}. 
This result is summarized below with its proof in the Appendix.
\begin{thm}\label{thm:fill}
Consider a geometric tree with parameters $\alpha > 0$ and $0< b \leq c$ 
as assumed in Theorem \ref{thm:geom} and let the rumor spread from 
source $v^*$ starting at time $0$. Define $\epsilon = t^{-1/2+\delta}$ 
for any small $0< \delta < 0.1$. Let $G(t)$ be the set of all rumor
infected nodes in the tree at time $t$. Let ${\mathcal G}_t$ be the
set of all sub-trees rooted at $v^*$ (rumor graphs) such that all nodes
within distance $t(1-\epsilon)$ from the $v^*$ are in the tree and but
no node beyond distance $t(1+\epsilon)$ from $v^*$ beyond to the tree.  
Then%
\begin{align*}
\mathbf P(G_t \in \mathcal G_t) & = 1 - O\big(e^{-t^\delta}\big) \\
                                &\stackrel{t\to\infty}{\longrightarrow} 1.
\end{align*}
\end{thm}
Define $\mathcal E_t$ as the event that $G_t \in \mathcal G_t$. Under
event $\mathcal E_t$, consider the sizes of the sub-trees $N_i(t)$ for $1\leq i\leq d^*$.
Due to the polynomial growth condition and $\mathcal  E_t$, we obtain the following
bounds on each $N_i(t)$ for all $1\leq i\leq d^*$:
\begin{align*}
\sum_{r=1}^{t(1-\epsilon)-1} b r^\alpha & \leq N_i(t) \\
                                      & \leq \sum_{r=1}^{t(1+\epsilon)-1} c r^\alpha.
\end{align*}
Now bounding the summations by Reimann's integrals, we have
\begin{align*}
\int_0^{L-1} r^\alpha dr & \leq \sum_{r=1}^L r^\alpha ~\leq~\int_0^{L+1} r^\alpha dr. 
\end{align*}
Therefore, it follows that under event $\mathcal E_t$, for all $1\leq i\leq d^*$
\begin{align*}
\frac{b}{1+\alpha} \big(t(1-\epsilon)-2\big)^{\alpha+1} & \leq N_i(t) ~\leq~ \frac{c}{1+\alpha}  \big(t(1+\epsilon)\big)^{\alpha+1}. 
\end{align*}
In the most `unbalanced' situation, $d^*-1$ of these sub-trees 
have minimal size $N_{\text{min}}(t)$ and the remaining one
sub-tree has size $N_{\text{max}}(t)$ where 
\begin{align*}
N_{\text{min}}(t) & = \frac{b}{1+\alpha} \big(t(1-\epsilon)-2\big)^{\alpha+1}, \\
N_{\text{max}}(t) & = \frac{c}{1+\alpha}  \big(t(1+\epsilon)\big)^{\alpha+1}.
\end{align*}
Since by assumption $c < b (d^*-1)$, there exists $\gamma > 0$ so that 
$c < (1+\gamma) b(d^*-1)$. Therefore, for choice of $\epsilon = t^{-1/2 + \delta}$ for
some $\delta \in (0,0.1)$, we have
\begin{align*}
\frac{(d^*-1)N_{\text{min}}(t)}{N_{\text{max}}(t)} &= \frac{b (d^*-1)}{c} \Big(\frac{t - t^{\frac{1}{2} +\delta} - 2}{t + t^{\frac{1}{2} + \delta}}\Big)^{\alpha+1} \\
& \stackrel{(i)}{>} \frac{1}{1+\gamma} \Big(\frac{1-t^{-\frac{1}{2} + \delta} - \frac{2}{t}}{1 + t^{-\frac{1}{2}+\delta}}\Big)^{\alpha+1} \\
& > 1,
\end{align*}
for $t$ large enough since the second term in inequality (i) 
goes to $1$ as $t\to\infty$. From this, it immediately follows
that under event $\mathcal E_t$ for $t$ large enough
\[ \max_{1\leq i\leq d^*} N_i(t) < \frac{1}{2} \sum_{i=1}^{d^*} N_i(t).\]
Therefore, by Proposition \ref{prop:rc} it follows that the rumor
center is unique and equals $v^*$. Therefore, for $t$ large
enough $\mathcal E_t \subset \mathcal C_t$. From above and Theorem \ref{thm:fill}
\begin{align*}
\liminf_{t} \P\big({\mathcal C}_t\big) & \geq \lim_{t} \P\big({\mathcal E}_t\big) \\
                                       & = 1.
\end{align*}

\section{Conclusion and Future Work}\label{sec:conclusion}
This paper has provided, to the best of the authors' knowledge, the first systematic study of the problem of finding rumor sources in networks.  Using the well known SIR model, we constructed an estimator for the rumor source in regular trees, general trees, and general graphs.  We defined the ML estimator for a regular tree to be a new notion of network centrality which we called rumor centrality and used this as the basis for estimators for general trees and general graphs. 

We analyzed the asymptotic behavior of the rumor source estimator for regular trees and geometric trees.    For linear graphs, it was shown that the detection probability goes to 0 as the network grows in size.  However, for trees which grew faster than lines, it was shown that there was always non-trivial detection probability.  This analysis highlighted the different techniques which must be used for networks with expansion versus those with only polynomial growth. Simulations performed on synthetic graphs agreed with these tree results and also demonstrated that the general graph estimator performed well in different network topologies, both synthetic (small-world, scale-free) and real (AS, power grid).  

On trees, we showed that the rumor center is equivalent to the distance center.  However, these were not equivalent in a general network.  Also, it was seen that in networks which are not tree-like, rumor centrality is a better rumor source estimator than distance centrality.

The next step of this work would be to better understand the effect of the BFS heuristic on the estimation error and under what precise conditions it improves or degrades performance.  Another future direction would be to generalize the estimator to networks with a heterogeneous rumor spreading rate.
\section{Appendix A: Proof of Theorem \ref{thm:fill}}\label{app:fill}

We recall that Theorem \ref{thm:fill} stated that the rumor graph on a geometric tree is full up to a distance $t(1-\epsilon)$ and does not extend beyond $t(1+\epsilon)$, for $\epsilon = t^{-1/2+\delta}$ for some positive 
$\delta \in (0,0.1)$. To establish this, we shall use the following well 
known concentration property of the unit rate Poisson process. We 
provide its proof later for completeness. 
\begin{thm}\label{thm:poisson}
Consider a unit rate Poisson process $P(\cdot)$ with rate 1. Then
there exists a constant $C > 0$ so that for any $\gamma \in (0,0.25)$, 
\begin{equation*}
	\mathbf{P}\big( \big| P(t) - t \big| \geq \gamma t \big) \leq 2 e^{-\frac{1}{4}t\gamma^2}.
\end{equation*}
\end{thm}
Now we use Theorem \ref{thm:poisson} to establish Theorem \ref{thm:fill}. Recall that
the spreading time along each edge is an independent and identically distributed exponential
random variable with parameter $1$. Now the underlying network graph is a tree. Therefore
for any node $v$ at distance $r$ from source node $v^*$, there is a unique
path (of length $r$) connecting $v$ and $v^*$. Then, the spread of the rumor
along this path can be thought of as a unit rate Poisson process, say $P(t)$, 
and node $v$ is infected by time $t$ if and only if $P(t) \geq r$. Therefore,
from Theorem \ref{thm:poisson} it follows that for any node $v$ that is
at distance $t(1-\epsilon)$ for $\epsilon = t^{-\frac{1}{2} + \delta}$ for some
$\delta \in (0,0.1)$, 
\begin{align*}
\P\big(v \text{~is not rumor infected}\big) & \leq 2 e^{-\frac{1}{4}t\epsilon^2} \\
                                            & = 2 e^{-\frac{1}{4}t^{2\delta}}. 
\end{align*}
Now the number of such nodes at distance $t(1-\epsilon)$ from $v^*$ is 
at most $O(t^{1+\alpha})$ (follows from arguments similar to those in
the proof of Theorem \ref{thm:geom}). Therefore, by an application of union
bound it follows that 
\begin{align*}
& \P\big(\text{a node at distance $t(1-\epsilon)$ from $v^*$ isn't infected}\big) \\
& \qquad = O\Big(t^{\alpha+1} e^{-\frac{1}{4}t^{2\delta}}\Big) \\
& \qquad = O\Big(e^{-t^\delta}\Big).
\end{align*}
Using similar argument and another 
application of Theorem \ref{thm:poisson}, it can be argued that
\begin{align*}
& \P\big(\text{a node at distance $t(1+\epsilon)$ from $v^*$ is infected}\big) \\
& \qquad = O\Big(e^{-t^{\delta}}\Big). 
\end{align*}
Since the rumor is a `spreading' process, if all nodes at distance $r$ from $v^*$ are
infected, then so are all nodes at distance $r' < r$ from $v^*$; if all nodes
at distance $r$ from $v^*$ are not infected then so are all nodes at distance $r' > r$ 
from $v^*$. Therefore, it follows that with probability $1-O(e^{-t^\delta})$, 
all nodes at distance up to $t(1-\epsilon)$ from $v^*$ are infected and all
nodes beyond distance $t(1+\epsilon)$ from $v^*$ are not infected. This
completes the proof of Theorem \ref{thm:fill}.

\section{Appendix B: Proof of Theorem \ref{thm:poisson}}\label{app:poisson}

We wish to prove bounds on the probability of $P(t) \leq t(1-\gamma)$ and
$P(t) \geq t (1+\gamma)$ for a unit rate Poisson process $P(\cdot)$. To
that end, for $\theta > 0$ it follows that 
\begin{align*}
	\P\big(P(t)\leq t(1-\gamma)\big) & = \P\big(-\theta P(t)\geq - \theta t(1-\gamma) \big) \\
	& = \P\big(e^{-\theta P(t)} \geq e^{-\theta t(1-\gamma)}\big) \\
	& \leq e^{\theta t(1-\gamma)} \E\big[e^{-\theta P(t)}\big] \\
	& = e^{\theta t(1-\gamma)} e^{t (e^{-\theta} -1)}, 
\end{align*}
where the last equality follows from the fact that $P(t)$ is a Poisson random
variable with parameter $t$. That is, 
\begin{align*}
	\P\big(P(t)\leq t(1-\gamma)\big) & \leq \inf_{\theta > 0} e^{\theta t(1-\gamma) + t e^{-\theta} - t}. 
\end{align*}
The minimal value of the exponent in the right hand side above is achieved for
value of $\theta = -\log (1-\gamma)$. For this value of $\theta$, using
the fact that $\gamma \in (0,0.25)$ and the inequality $\log (1-\gamma) \geq -\gamma - 3\gamma^2/4$ for
$\gamma < 1/3$, it follows that 
\begin{align*}
	\P\big(P(t)\leq t(1-\gamma)\big) & \leq e^{-\frac{1}{4}t\gamma^2}. 
\end{align*}
Next, to establish the bound on the probability of $P(t) \geq t(1+\gamma)$, using
similar argument it follows that 
\begin{equation*}
	\mathbf{P}\big(P(t)\geq t(1+\gamma)\big) \leq \inf_{\theta > 0} e^{t\big(-\theta(1+\gamma) +(e^{\theta}-1)\big) }.
\end{equation*}
The right hand side is minimized for $\theta = \log(1+\gamma)$. Using 
$\log(1+\gamma) \geq \gamma - \gamma^2/2$ for $\gamma \leq 0.5$ it follows
that 
\begin{equation*}
	\mathbf{P}\big(P(t)\geq t(1+\gamma)\big) \leq e^{-\frac{1}{4}t\gamma^2}. 
\end{equation*}
This completes the proof of Theorem \ref{thm:poisson}.

%

\section*{Acknowledgment}

Devavrat Shah would like to acknowledge a stimulating 
conversation with David Gamarnik and Andrea Montanari 
at the Banff International Research Station (BIRS) in 
the Summer of 2008 that seeded this work and would like
to thank the program at BIRS. Authors would like to 
acknowledge that this work was supported in parts by 
the AFOSR complex networks program, NSF HSD Project, 
NSF EMT Project and Shell Graduate Student Fellowship.


\begin{thebibliography}{10}


\bibitem{ref:sir}
N.~T.~J. Bailey.
\newblock {\em The Mathematical Theory of Infectious Diseases and its
  Applications}.
\newblock Griffin, London, 1975.

\bibitem{ref:epidemic-sw}
C.~Moore and M.~E.~J. Newman.
\newblock Epidemics and percolation in in small-world networks.
\newblock {\em Phys. Rev. E}, 61:5678--5682, 2000.

\bibitem{ref:epidemic-sf}
R.~Pastor-Satorras and A.~Vespignani.
\newblock Epidemic spreading in scale-free networks.
\newblock {\em Phys. Rev. Lett.}, 86:3200--3203, 2001.

\bibitem{ref:newman}
M.~E.~J. Newman.
\newblock The spread of epidemic disease on networks.
\newblock {\em Phys. Rev. E}, 66:016128, 2002.


\bibitem{ref:topology}
A.~Ganesh, L.~Massoulie, and D.~Towsley.
\newblock The effect of network topology on the spread of epidemics.
\newblock {\em Proc. 24th Annual Joint Conference of the IEEE Computer and
  Communications Societies (INFOCOM)}, 2:1455--1466, 2005.
  




\bibitem{ref:mcmc1} N. Demiris and P. D. O'Neill, ``Bayesian inference for epidemics with two levels of mixing'',  \textit{Scandinavian J. of Statistics}, vol. 32, pp. 265 - 280 (2005).
                    
\bibitem{ref:mcmc2}
G.~Streftaris and G.~J. Gibson.
\newblock Statistical inference for stochastic epidemic models.
\newblock {\em Proc. 17th international Workshop on Statistical Modeling},
  pages 609--616, 2002.
                      
\bibitem{ref:mcmc3}  P. D. O'Neill, ``A tutorial introduction to Bayesian inference for stochastic epidemic models using Markov chain Monte Carlo methods,'' \textit{Mathematical Biosciences} vol. 180, pp. 103-114. (2002).

\bibitem{ref:mcmc4} N. Demiris and P. D.  O'Neill, ``Bayesian inference for stochastic multitype epidemics in structured populations via random graphs,'' \textit{J. Roy. Statist. Soc. B}, vol. 67, pp. 731-745. (2005).


\bibitem{ref:nhpp}
H.~Okamura, K.~Tateishi, and T.~Doshi.
\newblock Statistical inference of computer virus propagation using
  non-homogeneous poisson processes.
\newblock {\em Proc. 18th IEEE International Symposium on Software
  Reliability}, 5:149 -- 158, 2007.




\bibitem{ref:evans} W. Evans, C. Kenyon, Y. Peres, and L. Schulman,  ``Broadcasting on trees and the Ising model'', \textit{Ann. Appl. Prob.}, vol. 10, pp. 410-433. (2000).  


\bibitem{ref:mossel} E. Mossel, ``Reconstruction on trees: beating the second eigenvalue'', \textit{Ann. Appl. Prob.}, vol. 11, pp. 285-300. (2001). 


\bibitem{ref:montanari}  A. Gerschenfeld and A. Montanari, ``Reconstruction for models on random graphs,'' \textit{Proc. 48'th IEEE Symp. Found. Comp. Sci.} pp. 194-204 (2007). 

\bibitem{ref:dc}
G.~Sabidussi.
\newblock The centrality index of a graph.
\newblock {\em Psychometrika}, 31:581--603, 1966.

\bibitem{ref:bc}
L.~C. Freeman.
\newblock A set of measure of centrality based on betweenness.
\newblock {\em Sociometry}, 40:35--41, 1977.


\bibitem{ref:shape1} J. M. Hammersley and D.J.A. Welsh, ``First-passage percolation, subadditive processes, stochastic networks, and generalized renewal theory,'' \textit{Bernoulli-Bayes-Laplace Anniversary Volume}, Springer, Berlin (1965).

\bibitem{ref:shape2} R. Smythe and J. C. Wierman, \textit{First Passage Percolation on the Square Lattice, Lecture Notes in Math}, Springer, Berlin (1978).

\bibitem{ref:shape3} J. T. Cox and R. Durrett, ``Some limit theorems for percolation processes with necessary and sufficient conditions,'' \textit{Ann. Appl. Prob.}, vol. 9, pp. 583-603. (1981).

\bibitem{ref:shape4} H. Kesten, \textit{Aspects of First Passage Percolation.  Lecture Notes in Math}, vol. 1180, pp. 125-264. Springer, Berlin (1986).

\bibitem{ref:shape5} H. Kesten, ``Percolation theory and first-passage percolation,'' \textit{Ann. Appl. Prob.}, vol. 15, pp. 1231-1271. (1987).

\bibitem{ref:shape6} H. Kesten, ``On the speed of convergence in first-passage percolation,'' \textit{Ann. Appl. Prob.}, vol. 3, pp. 296-338. (1993).


\bibitem{ref:posetalg} M. D. Atkinson, ``On computing the number of linear extensions of a tree,'' \textit{Order}, vol. 7, pp. 23--25. (1990).

\bibitem{ref:posetcomplex} G. Brightwell and P. Winkler, ``Counting linear extensions is \#P-complete,'' \textit{JSTOC '91: Proceedings of the twenty-third annual ACM symposium on Theory of computing}, pp. 175--181. (1991).





\bibitem{ref:sw}
D.~J. Watts and S.~Strogatz.
\newblock Collective dynamics of `small-world' networks.
\newblock {\em Nature}, 393:440--442, 1998.

\bibitem{ref:sf}
A.~Barabasi and R.~Albert.
\newblock Emergence of scaling in random networks.
\newblock {\em Science}, 286:509--512, 1999.



\bibitem{ref:AS}
The {CAIDA AS} relationships dataset.
\newblock {\em http://www.caida.org/data/active/as-relationships/}, August
  30th, 2009.
  
\bibitem{BT09}
D. Betsekas and J. N. Tsitsiklis.
\newblock Probabilistic System Analysis. 
\newblock Aetna publication, 2nd edition, 2009. 

\end{thebibliography}
\end{document}